%% file: example_paper.tex
\documentclass{article}

\usepackage{microtype}
\usepackage{graphicx}
\usepackage{subcaption}
\usepackage{booktabs} 
\usepackage{multirow}

\usepackage{hyperref}

\usepackage[preprint]{icml2026}

\usepackage{amsmath}
\usepackage{amssymb}
\usepackage{mathtools}
\usepackage{amsthm}

\usepackage[capitalize,noabbrev]{cleveref}

\theoremstyle{plain}
\newtheorem{theorem}{Theorem}[section]
\newtheorem{proposition}[theorem]{Proposition}

\theoremstyle{definition}

\theoremstyle{remark}

\usepackage[textsize=tiny]{todonotes}

\icmltitlerunning{CoCA: Step-Level Reward for Free in RL-Finetuned T2I Diffusion Models}
\begin{document}

\twocolumn[
  \icmltitle{CoCA: Step-Level Reward for Free in RL-Finetuned \\ T2I Diffusion Models}
  \icmlsetsymbol{equal}{*}

  \begin{icmlauthorlist}
    \icmlauthor{Xinyao Liao}{hust,nus}
    \icmlauthor{Wei Wei}{hust}
    \icmlauthor{Xiaoye Qu}{hust}
    \icmlauthor{Qiyuan He}{nus}
    \icmlauthor{Angela Yao}{nus}
    \icmlauthor{Yu Cheng}{cuhk}
  \end{icmlauthorlist}

  \icmlaffiliation{hust}{HUST}
  \icmlaffiliation{nus}{NUS}
  \icmlaffiliation{cuhk}{CUHK}
  \icmlcorrespondingauthor{Wei Wei}{weiw@hust.edu.cn}

  \icmlkeywords{Diffusion Models, Reinforcement Learning, Credit Assignment, Text-to-Image Generation}

  \vskip 0.3in
]

\printAffiliationsAndNotice{} 

\input{sec/0_abstract}    
\input{sec/1_intro}
\input{sec/2_related_work}
\input{sec/3_prelimilaries}
\input{sec/4_method}
\input{sec/5_experiments}
\input{sec/6_conclusion}

\section*{Impact Statement}

This work studies credit assignment for reinforcement-learning fine-tuning of text-to-image diffusion models.
By providing denser learning signals derived from terminal feedback, our method may reduce the compute and data required to align generative models to human preference, which can lower the barrier to beneficial applications such as content creation and assistive design.

At the same time, improved fine-tuning efficiency can also facilitate misuse (e.g., generating deceptive or harmful imagery) and may amplify biases inherited from pretrained models and reward models.

\bibliography{example_paper}
\bibliographystyle{icml2026}

\input{sec/7_suppl}

\end{document}

%% file: sec/0_abstract.tex
\begin{abstract}
State-of-the-art text-to-image diffusion models often leverage reinforcement learning for fine-tuning, but they suffer from severe reward sparsity due to the lack of reliable intermediate evaluation. Since existing methods typically receive only a single terminal reward, they fail to identify which specific denoising steps meaningfully contribute to final image quality. To address this, we propose CoCA, a simple and general contribution-based credit assignment framework that redistributes the terminal reward across the denoising trajectory. CoCA estimates each step's contribution by measuring the latent progression toward the clean image, enabling dense and informative rewards without requiring auxiliary networks. Our approach integrates seamlessly with various policy-gradient algorithms, such as REINFORCE and GRPO, through reward shaping. Empirical results demonstrate that CoCA improves optimization efficiency by 25 to 100\%, while preserving the original objective under policy optimization.

\end{abstract}

%% file: sec/1_intro.tex
\section{Introduction}
\label{sec:intro}
\begin{figure}
    \centering
    \includegraphics[width=0.8\linewidth]{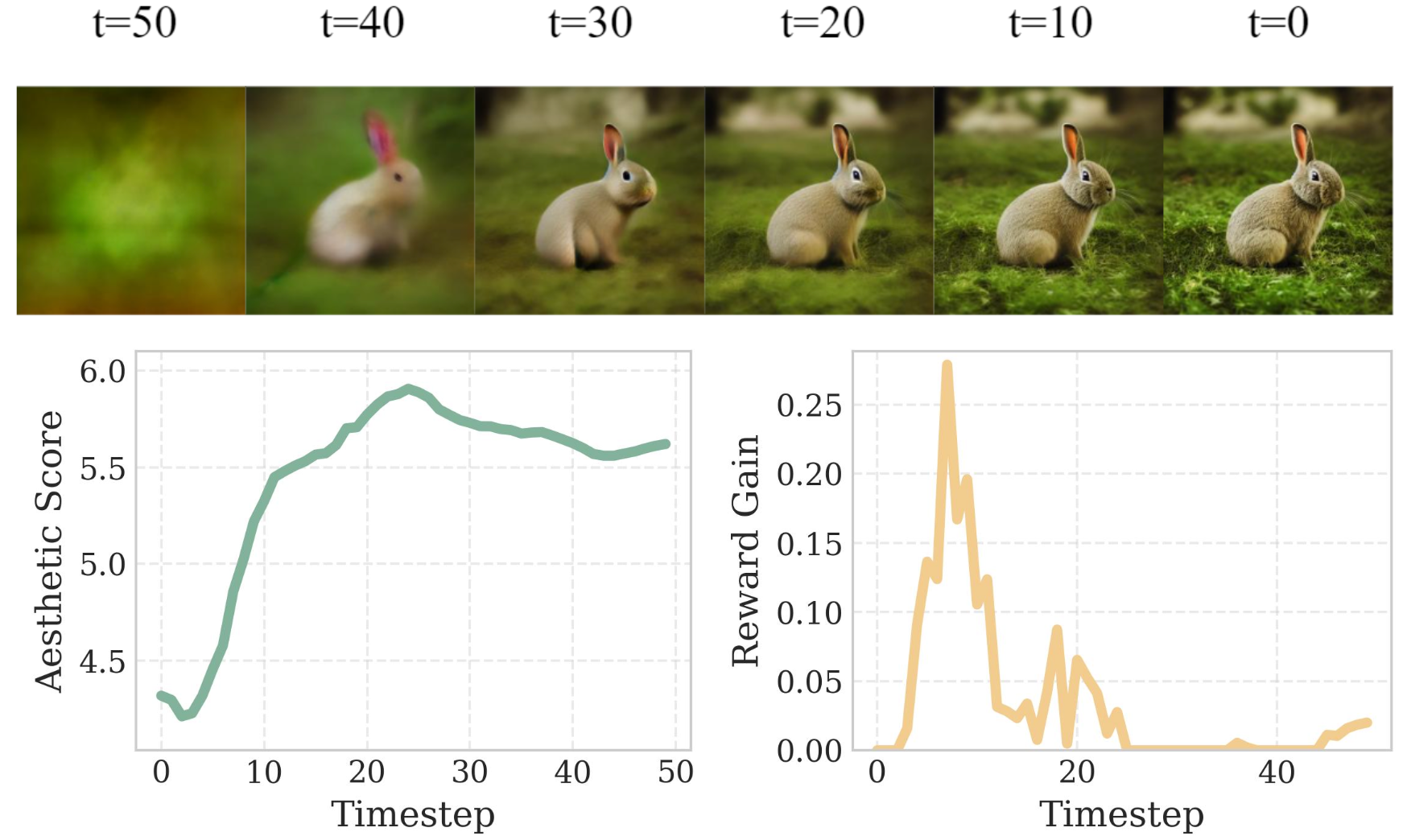}
    \caption{
    \textbf{Reward shaping challenge in diffusion denoising.} (1) Top: intermediate predicted images from $t{=}50$ to $t{=}0$ show the global structure restored in the early stage and fine-grained details refined in later steps. (2) Bottom-left: the aesthetic score increases as denoising proceeds but plateaus in later steps. (3) Bottom-right: the step-wise reward gain rapidly diminishes over time, indicating that later denoising steps contribute far less to final image quality.
    \vspace{-3em}
    }
    \label{fig:intro-obs}
\end{figure}

\begin{figure*}
    \centering
    \includegraphics[width=0.7\linewidth]{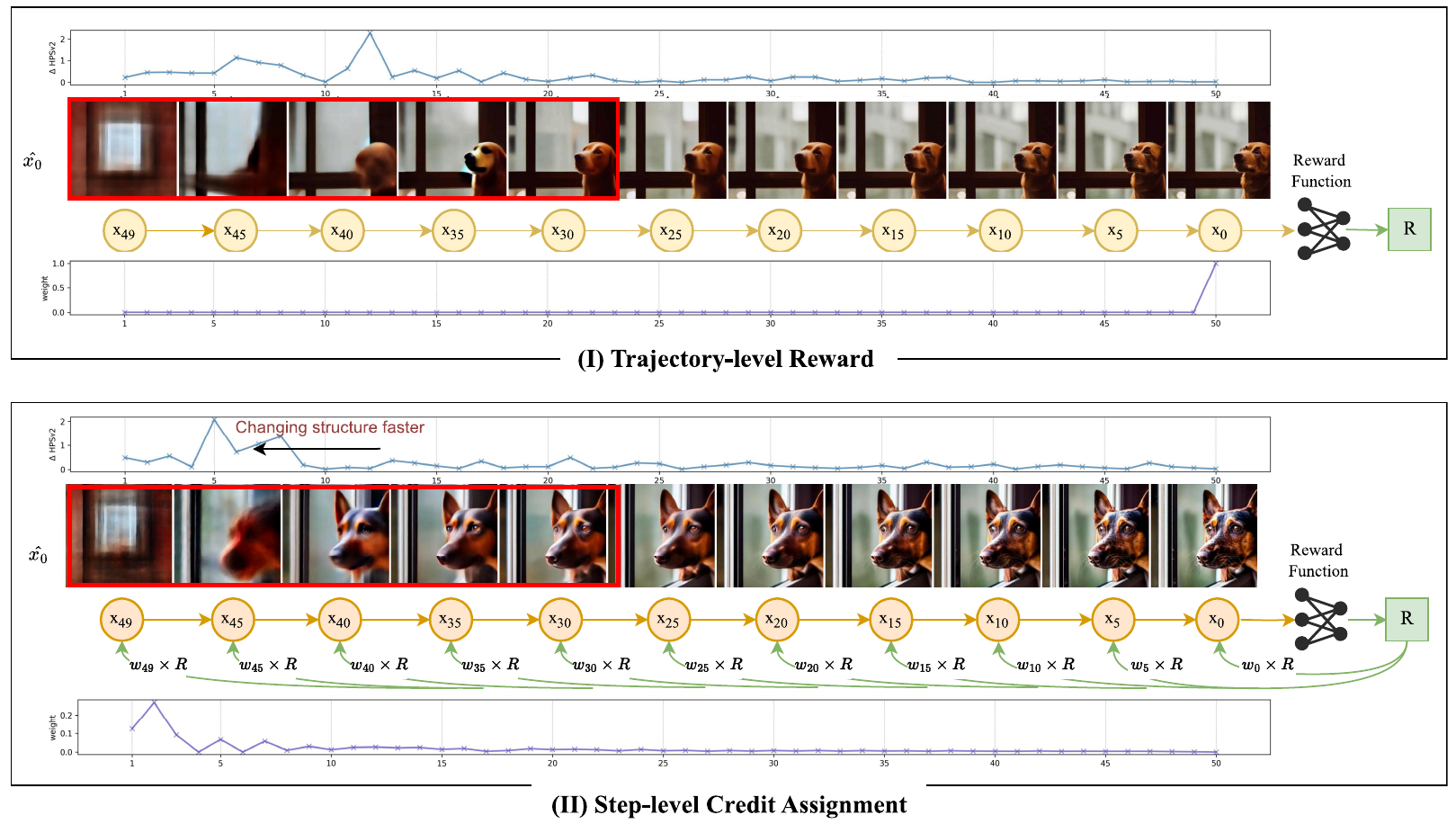}
    \caption{\textbf{Overview of trajectory-level reward and step-level credit assignment.} (I) Trajectory-level sparse rewards treat all timesteps uniformly, offering no guidance on where improvements occur.
    (II) CoCA converts this scalar reward into dense, step-aware credit by weighting each timestep according to its measured contribution. Visualizing the reward gains ($\Delta$HPSv2~\citep{hps2023}) across timesteps shows that CoCA accelerates the formation of coherent global structure earlier in the denoising process.}
  \vspace{-1.5em}
  \label{fig:intro-main}
\end{figure*}

Diffusion models~\citep{diffusion2015, ddpm2020, ddim2020,tian2025extrapolating} have emerged as the dominant paradigm in image generation, offering high image fidelity and easy scalability. 
Recent advances in text-to-image (T2I) diffusion models~\citep{sd2022, dalle32023, flux} have greatly expanded the realism and expressiveness of visual content generation.
However, despite strong pretrained capabilities, their outputs may not fully align with human preference.  In particular, characteristics like aesthetic appeal and subtle perceptual qualities tend to fall outside strict semantic correctness.  This gap has motivated a growing interest in modeling human preference~\cite{hps2023,imagereward2023} and post-training diffusion models for alignment~\cite{dpo2023,ddpo2023,dancegrpo2025}.

To address this alignment gap, recent work explores post-training strategies that directly optimize diffusion models using human feedback. Methods based on differentiable reward models~\citep{imagereward2023,unifl} enable end-to-end optimization, though they require rewards to be differentiable and incur substantial memory overhead. DPO-style preference optimization~\citep{dpo2023,difdpo2024,dpoppo2024} offers a simpler training pipeline but typically yields only modest gains in visual fidelity. In contrast, reinforcement learning (RL)~\cite{ddpo2023,dpok2023,dancegrpo2025} has emerged as a promising paradigm, as it does not require differentiable rewards and can be optimized with continuous signals.

Recent work views diffusion as a policy model~\citep{ddpo2023,dpok2023},  formulating iterative denoising as a Markov decision process. Under this formulation, policy-gradient algorithms~\citep{policygradient1999,ppo2017,grpo} can be applied to optimize the model with reward functions derived from human feedback.
These approaches demonstrate improved alignment with prompts and better quality perceived by humans. 

However, the denoising trajectory of generating images typically shows diminishing marginal improvement in reconstruction quality at later timesteps~\cite{sd2022, simada2024, bluenoise2024, boost2024, fresca}. As shown in  Figure~\ref{fig:intro-obs}, early timesteps play a decisive role in determining the global structure of the image, while later steps contribute mainly to fine-grained details.  However, existing policy gradient methods typically apply a sparse reward signal only at the end point and update the policy equally across all timesteps, as shown in Figure~\ref{fig:intro-main}. 

This leads to a mismatch between the similar magnitude of policy updates and the unequal importance of different timesteps. Consequently, existing RL-based methods suffer from suboptimal sample efficiency and limited precision in attributing rewards to specific generative actions~\cite{hindsight2019,timecredit1984}. Recent efforts to address this mismatch apply fixed temporal discounting to prioritize early denoising steps~\cite{densereward2024}, or train auxiliary networks to predict step-wise critics or preferences~\cite{tdpo2024, spo2024, lpo2025}. Crucially, existing methods either impose a priori assumptions about step importance or require costly reward model expansion to mitigate reward sparsity. 

This work poses a fundamental question: \textit{Can we quantify the 
contribution of each denoising step to the final image quality?}  To address this, we propose a novel framework that enables \textbf{\underline{Co}}ntribution-based \textbf{\underline{C}}redit \textbf{\underline{A}}ssignment (\textbf{CoCA}) for diffusion-based text-to-image generation as shown in Figure~\ref{fig:intro-main} (II).
Specifically, for each denoising step, we track similarity progression between the intermediate latent and the realized terminal latent in embedding space, yielding an outcome-aligned latent-progress proxy.
We use this proxy to redistribute the terminal reward into denser learning signals, keeping the optimization target aligned with the same terminal objective rather than introducing any additional per-step objective. Then, we convert the sparse trajectory-level reward into informative step-level rewards by redistributing the terminal reward according to the estimated step contributions, without training additional networks.

To further stabilize training, we normalize rewards once before redistribution to maintain per-prompt ranking, and once after redistribution to keep reward scales consistent across timesteps and reduce variance during optimization. 
Theoretically, we show that this reward redistribution does not bias the expected policy gradient, and thus preserves the policy-gradient stationary points of the original objective.

Experiments across four reward functions based on DDPO~\citep{ddpo2023} demonstrate that our framework achieves 25\%-100\% improvement of sample efficiency than trajectory-level reward baselines~\cite{ddpo2023,dancegrpo2025} and step-level reward baselines~\cite{tdpo2024}. We also apply CoCA to DanceGRPO~\cite{dancegrpo2025}, the state-of-the-art GRPO~\cite{grpo} framework in T2I diffusion.  The quantitative study shows that our method exhibits better prompt alignment and improves the generalization capability on both unseen rewards and unseen prompts, all without sacrificing computational simplicity, demonstrating rapidly changing the global layout. 

To summarize, our contributions are as follows: (1) We introduce CoCA, a novel credit assignment framework to deal with the mismatch between equal policy updates and the varying impact of steps caused by reward sparsity, without training additional networks. 
(2) We empirically observe that timestep-level latent-progress patterns vary substantially across prompts/trajectories (e.g., Fig.~\ref{fig:power_prompt}), which helps explain why fixed temporal weighting schedules fail to generalize and motivates adaptive credit assignment. 
(3) Comprehensive experiments across human preference score (Aesthetic~\cite{aesthetic2022}, ImageReward~\cite{imagereward2023}, HPSv2~\cite{hps2023}, PickScore~\cite{pick2023}) demonstrate superior sample efficiency and generalization capabilities compared to recent trajectory/step-level methods on cross-rewards and unseen prompts.

%% file: sec/2_related_work.tex
\section{Related Work}
\label{sec:relatedwork}

\subsection{Human Preference Fine-Tuning}
Human preference-guided fine-tuning of text-to-image diffusion models generally follows three directions. 
Methods based on differentiable reward models~\cite{imagereward2023,raft2023,drtune2024} 
enable gradient-based updates but restrict reward design and introduce substantial memory overhead. 
DPO-style preference optimization~\cite{difdpo2024,d3po2024,spo2024} removes the need for explicit reward modeling, 
yet offers relatively coarse training signals and typically yields incremental gains in perceptual quality.

Reinforcement learning approaches~\cite{ddpo2023,dpok2023,dancegrpo2025} 
formulate the denoising process as a Markov decision process and optimize arbitrary reward functions via policy gradients. 
However, prior policy-gradient methods~\citep{ddpo2023,dpok2023} have been observed to exhibit instability when scaling beyond relatively small datasets. 
More recent variants, GRPO-based diffusion fine-tuning~\cite{dancegrpo2025}, demonstrate improved stability and scalability over large prompt sets.

However, existing RL-based diffusion methods predominantly rely on trajectory-level sparse rewards, which provide limited guidance for the multi-step denoising process~\citep{sd2022, simada2024, bluenoise2024, boost2024, fresca}. 
We study credit assignment in RL-fine-tuned diffusion models~\cite{stepai1961,timecredit1984,uniform2020,hindsight2019}, aiming to better distribute terminal rewards across denoising steps. Our formulation is algorithm-agnostic and can be integrated into various RL frameworks.

\subsection{Dense Reward for RL Fine-tuned Models}
Sparse rewards pose a significant challenge in reinforcement learning due to the difficulty of credit assignment over long horizons~\cite{shaping1999}. Prior work constructs denser supervision to improve sample efficiency and stability, e.g., using large language models (LLMs) to generate executable reward functions~\cite{text2reward2024} or learning token-level rewards for RLHF~\cite{dpoppo2024, r3hf2024, abc2024}. For diffusion models, sparse terminal feedback has motivated timestep-aware signals via heuristic temporal weighting~\cite{densereward2024}, step-aware preference optimization~\cite{spo2024}, and learned critics~\cite{tdpo2024}. More recently, LPO~\cite{lpo2025} learns a noise-aware latent reward model to provide data-driven step-level preference signals. Unlike methods~\cite{lpo2025} learns a data-driven step-level reward model with additional parameters and supervision, CoCA operates purely as a reward redistribution mechanism conditioned on realized trajectories, requiring no auxiliary training or reward annotations.

%% file: sec/3_prelimilaries.tex
\section{Preliminaries}
\label{sec:prelimilaries}


\subsection{Diffusion Models}

\paragraph{Formulations.}
This work considers a denoising diffusion probabilistic model (DDPMs)~\cite{ddpm2020}, which samples high-quality results in visual generation scenarios. Given samples from a data distribution $q(x_0)$, DDPMs aim to approximate $q(x_0)$ by using a latent variable model $p_{\theta}(x_0) \mathrel{\coloneqq} \int p_{\theta}(x_{0:T}) \, dx_{1:T}$. In this model, latent variables $x_1, ..., x_T$ are sampled from the \textit{forward process} that defines an approximate posterior $q(x_{1:T}|x_0)$. The forward process is a Markov chain that adds Gaussian noise gradually according to a variance schedule $\{\beta_t \in (0,1)\}_{t=1}^T$ over $T$ timesteps:
\begin{equation}
    \begin{aligned}
      &q(x_{1:T}|x_0) \coloneqq \prod_{t=1}^T q(x_t|x_{t-1}),
       \\&q(x_t|x_{t-1}) \coloneqq \mathcal{N} (x_t; \sqrt{1-\beta_t}x_{t-1}, \beta_t\mathbf{I})
    \end{aligned}
\end{equation}

At each step $t$, $x_t$ is sampled as $x_t = \sqrt{\alpha_t}x_{t-1}+\sqrt{1-\alpha_t}\epsilon_t$, where $\epsilon_t \sim \mathcal{N}(\mathbf{0}, \mathbf{I})$ and $\alpha_t \coloneqq 1 - \beta_t$. By leveraging the closure property of normal distributions, $x_t$ can also be expressed as $x_t = \sqrt{\bar{\alpha}_t}x_0+\sqrt{1 - \bar{\alpha}_t}\bar{\epsilon}_t$, where $\bar{\epsilon}_t \sim \mathcal{N}(0,1)$ and $\bar{\alpha}_t \coloneqq \prod_{i=1}^{t}\alpha_i$. The \textit{reverse process} learns the reverse version of the Markov chain of diffusion process starting at $p(x_T)=\mathcal{N}(x_T;\mathbf{0}, \mathbf{I})$:
\begin{equation}
    \begin{aligned}
      &p_\theta(x_{0:T}) \coloneqq p(x_T)\prod_{t=1}^T p_\theta(x_{t-1}|x_t),
       \\&p_\theta(x_{t-1}|x_t) \coloneqq \mathcal{N}(x_{t-1};\mu_\theta(x_t,t), \sigma^2 \mathbf{I})
    \end{aligned}
\end{equation}

\paragraph{Application on T2I Generation.}
In T2I generation, the \emph{context} $c$ corresponds to the input text prompt, and the model aims to generate an image sample $x_0$ that is semantically consistent with that prompt. Diffusion models learn the conditional distribution $q(x_0 \mid c)$ by minimizing a variational bound on the negative log-likelihood $\mathbb{E}q[-\log p\theta(x_0 \mid c)]$, which can be expressed as:
\begin{equation}
    \mathcal{L} \coloneqq \mathbb{E}_q[\Sigma_{t=1}^T \mathbb{D}_{\text{KL}}(q(x_{t-1}|x_t, x_0) \parallel p_\theta(x_{t-1}|x_t, c))].
\end{equation}

Ho et al.~\cite{ddpm2020} choose the parameterization $\mu_\theta(x_t, t)=\frac{1}{\sqrt{\alpha_t}}(x_t-\frac{\beta_t}{\sqrt{1-\bar{\alpha}}_t}\epsilon_\theta(x_t,t))$ to predict the noise at each step. the optimization objective of each step can be simplified as: 
\begin{equation}
    \mathcal{L}_t \coloneqq \mathbb{E}_{x_0, \epsilon}[\| \epsilon - \epsilon_\theta(\sqrt{\bar{a}_t}x_0 + \sqrt{1-\bar{\alpha}_t}\epsilon,c,t)\|^2].
\end{equation}

Although this denoising objective enables strong generative capability, it does not explicitly optimize for human-centric visual qualities. 

\subsection{Reinforcement Learning}
\paragraph{RL Objectives.}
Following recent works, we consider treating diffusion models as a Markov decision process (MDP) to enable RL training. A standard finite-state MDP is defined by a tuple $(\mathcal{S}, \mathcal{A}, P, R,\gamma)$, where $\mathcal{S}$ represents the state space, $\mathcal{A}$ represents the action space, $P$ represents the transition kernel that maps the current state and action to the next state, $R$ represents the reward function, $\gamma$ represents the discount factor when comes to cumulative returns. The agent takes a sequence of actions following the discrete timestep schedule $t \in (0, 1, \dots, T)$. At each timestep $t$, an agent perceives the current state $s_t$ and takes the action of $a_t$ according to a policy $\pi(a_t|s_t)$. The agent produces a trajectory of states and actions $\tau \coloneqq (s_0, a_0, s_1, a_1, \dots, s_T)$. The RL objective is optimizing the policy to maximize the expected cumulative reward over trajectories as follows:
\begin{equation}
    \label{rl_obj}
    \mathcal{J}(\pi) \coloneqq \mathbb{E}_{\tau \sim p(\tau|\pi)}\left[\sum_{t=0}^{T-1}R(s_t, a_t)\right]
\end{equation}

\paragraph{Trajectory-level Reward and Policy Optimization.}

Let \( p_\theta(x_{0}|c) \) denote a T2I diffusion model, where \( c\sim p(c) \) represents the text prompt distribution, and \( r(x_0, c) \) is a reward obtained from the final step. 

We formalize the denoising process as a \( T \)-step Markov Decision Process (MDP) with the following components:
\begin{equation}
    \label{MDP}
    \begin{aligned}
    &s_t \coloneqq (x_{T-t}, c),\ a_t \coloneqq x_{T-t-1},\ P(s_{t+1}|s_t, a_t) \coloneqq (\delta_{c}, \delta_{a_{t}}), \ \\
    & \pi_\theta(a_t|s_t) \coloneqq p_\theta(x_{T-t-1}|x_{T-t},c),\ 
    P(s_0) \coloneqq (p(c), \mathcal{N}(0,\mathbf{I}))
    \\&R(s_t,a_t) \coloneqq 
        \begin{cases}
            r(x_0,c), & t=T-1 \\
            0, & t < T-1
        \end{cases}
    \end{aligned}
\end{equation}

\noindent
Here, $s_t$ and $a_t$ denote the state and action at timestep $t$, $P(s_0)$ is the initial state distribution, and the parameterized policy $\pi_\theta$ corresponds to the underlying diffusion model. 
The transition kernel $P$ is deterministic, as executing a denoising action uniquely determines the next state, which we express using the Dirac distribution $\delta_y$. 
Under trajectory-level rewards, each rollout receives a single score $r(s_0, c)$.

This MDP mirrors the reverse diffusion process: Starting from Gaussian noise $x_T$, the policy $\pi_\theta$ iteratively refines the latent state over $T$ steps to generate $x_0$. According to Eq. \eqref{rl_obj}, the RL objective of the trajectory-level $\mathcal{J}_{\text{TR}}$ can be written as maximizing the expected final reward:

\begin{equation}
    \label{tr_obj}
    \mathcal{J}_{\text{TR}}(\pi_\theta) \coloneqq \mathbb{E}_{\tau
    \sim \pi_\theta}\left[\sum_{t=0}^{T-1}R(s_t, a_t)\right] = \mathbb{E}_{\tau
    \sim \pi_\theta}\left[r(x_0, c)\right]
\end{equation}

Using the Monte-Carlo policy gradient, also known as REINFORCE~\cite{mcpg2004} , we derive the policy gradient:

\begin{equation}
\begin{aligned}
&\nabla_\theta \mathcal{J}_{\text{TR}}(\pi_\theta)=\nabla_\theta\mathbb{E}_{\tau
\sim \pi_\theta}\left[r(x_0,c)\right] \\
&= \mathbb{E}_{\tau
\sim \pi_\theta}\left[r(x_0,c)\sum_{t=1}^T\nabla_\theta\log p_\theta(x_{t-1}|x_t,c)\right]
\end{aligned}
\label{eq:policy}
\end{equation}

This sparse-reward formulation creates a credit assignment challenge: The gradient in Equation~\ref{eq:policy} equally weights all denoising steps through the coefficient $r(x_0,z)$, despite their varying impacts on the final image quality.

%% file: sec/4_method.tex
\section{Method}
\label{sec:method}

In this section, we begin by whether heuristic reward reweighting schedule can ease reward sparsity, showing that power-based early emphasis provides limited gains and fails to generalize across prompts (Sec.~\ref{ssec:heuristic}). This motivates CoCA, which measures each denoising step’s latent-space progress toward the clean image and redistributes the final reward proportionally based on this progress proxy. With window smoothing and reward normalization, CoCA produces stable step-level rewards while preserving the stationary points of the original objective under policy gradient optimization (Sec.~\ref{ssec:coca}).

\subsection{Heuristic Schedule Analysis}
\label{ssec:heuristic}

Before introducing our contribution-based approach, we first explore a simple \textit{heuristic weighting} scheme to redistribute the final reward across denoising steps. 
Previous works~\cite{ddpo2023, dpok2023} assign uniform credit across timesteps, despite earlier diffusion steps having greater influence on final image quality.
To capture this, we introduce a power-based weighting schedule:
\begin{equation}
    w_t = \frac{(T - t + 1)^p}{\sum_{k=1}^{T}(T - k + 1)^p},
\end{equation}
where \(p > 0\) controls the degree of early emphasis. 
Larger \(p\) allocates more reward to early denoising steps, while \(p=0\) corresponds to uniform weighting. 

\begin{figure}[t]
    \centering
    \includegraphics[width=.9\linewidth]{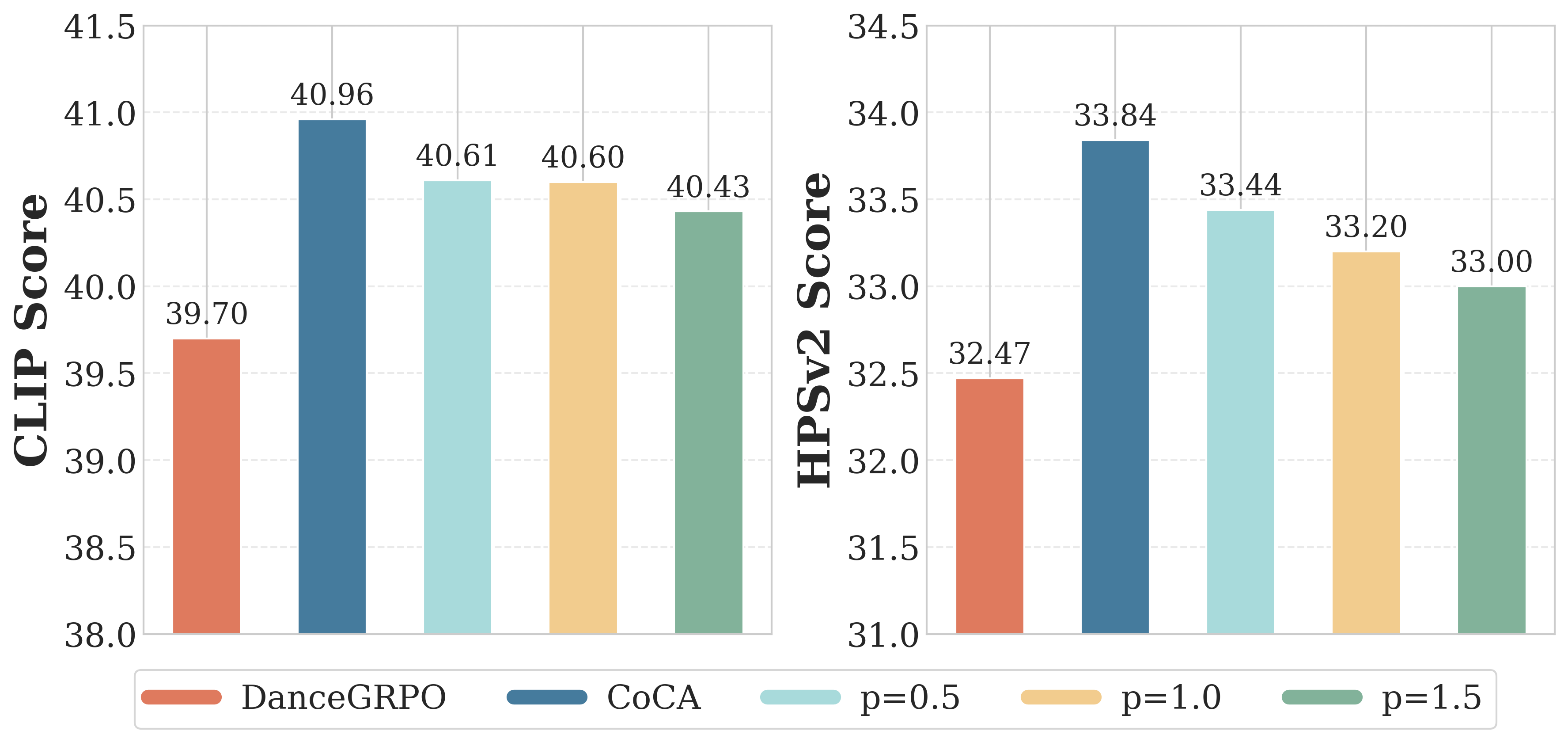}
    \caption{
    \textbf{Ablation of heuristic weight exponent \(p\) and comparison with CoCA under the DanceGRPO setting.}
    We compare heuristic weighting (\(p=0.5, 1.0, 1.5\)) with DanceGRPO~\citep{dancegrpo2025} and our CoCA using CLIP~\citep{clip2021} and HPSv2~\citep{hps2023}.
    Moderate early emphasis (\(p=0.5\)) offers slight gains, while CoCA adaptively learns timestep importance, achieving the best overall performance.
    }
    \label{fig:ablation_power_early}
    \vspace{-1.5em}
\end{figure}

We evaluate this heuristic under the DanceGRPO multi-reward setting as discussed in Section~\ref{ssec:imp}. 
As shown in Figure~\ref{fig:ablation_power_early}, moderate early emphasis (\(p=0.5\)) improves both CLIP~\citep{clip2021} and HPSv2~\citep{hps2023} scores compared to uniform weighting, while excessive emphasis (\(p=1.5\)) performances worse. 
Notably, our proposed CoCA surpasses all heuristic schedules. 
It suggests although fixed temporal priors can partially capture timestep importance, they fail to generalize across diverse prompts, highlighting the need for adaptive credit assignment.

\begin{figure}[t]
    \centering
    \begin{minipage}[b]{0.45\linewidth}
        \centering
        \includegraphics[width=\linewidth]{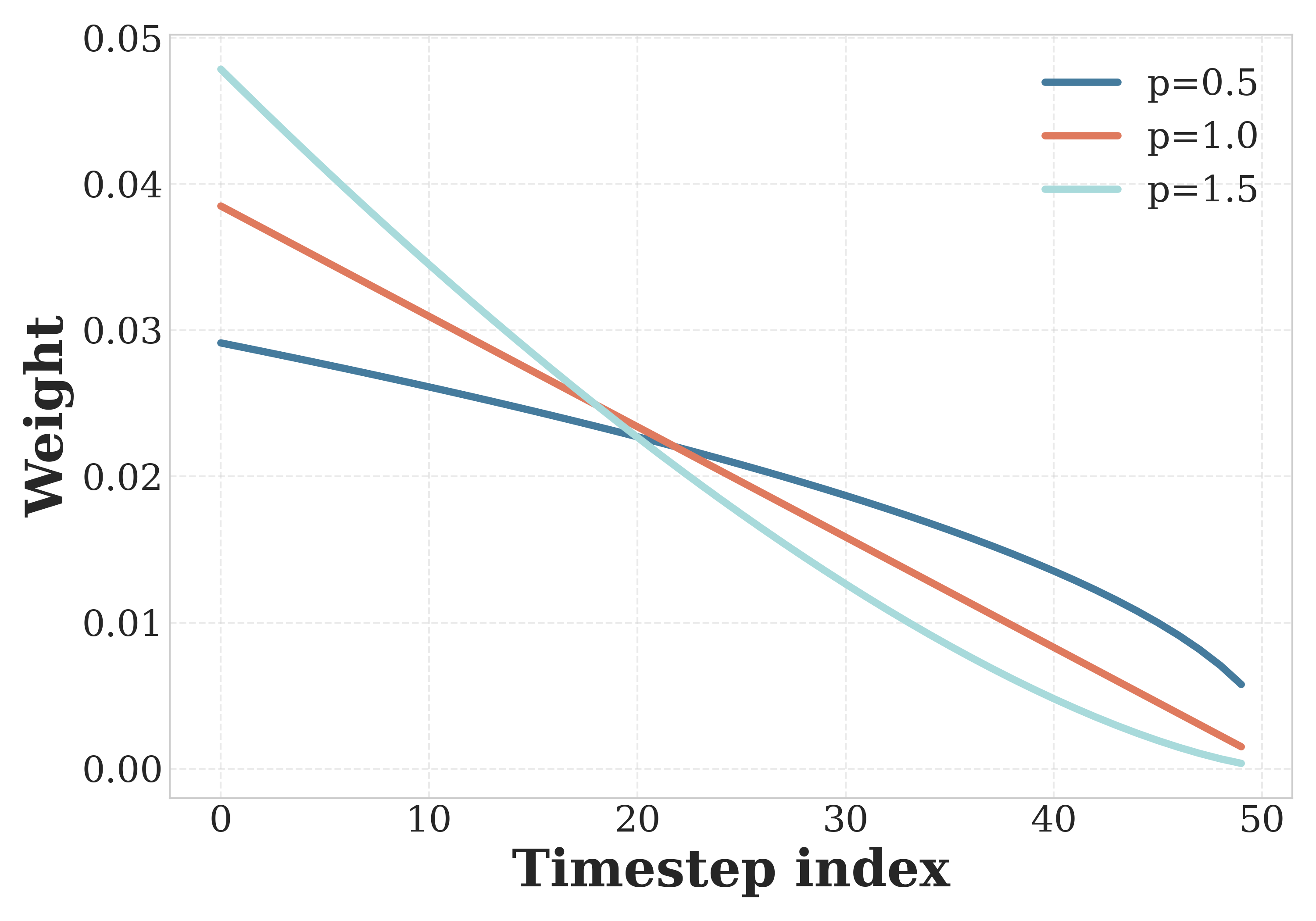}
        \caption*{(a)}
        \label{fig:power_early}
    \end{minipage}
    \hfill
    \begin{minipage}[b]{0.45\linewidth}
        \centering
        \includegraphics[width=\linewidth]{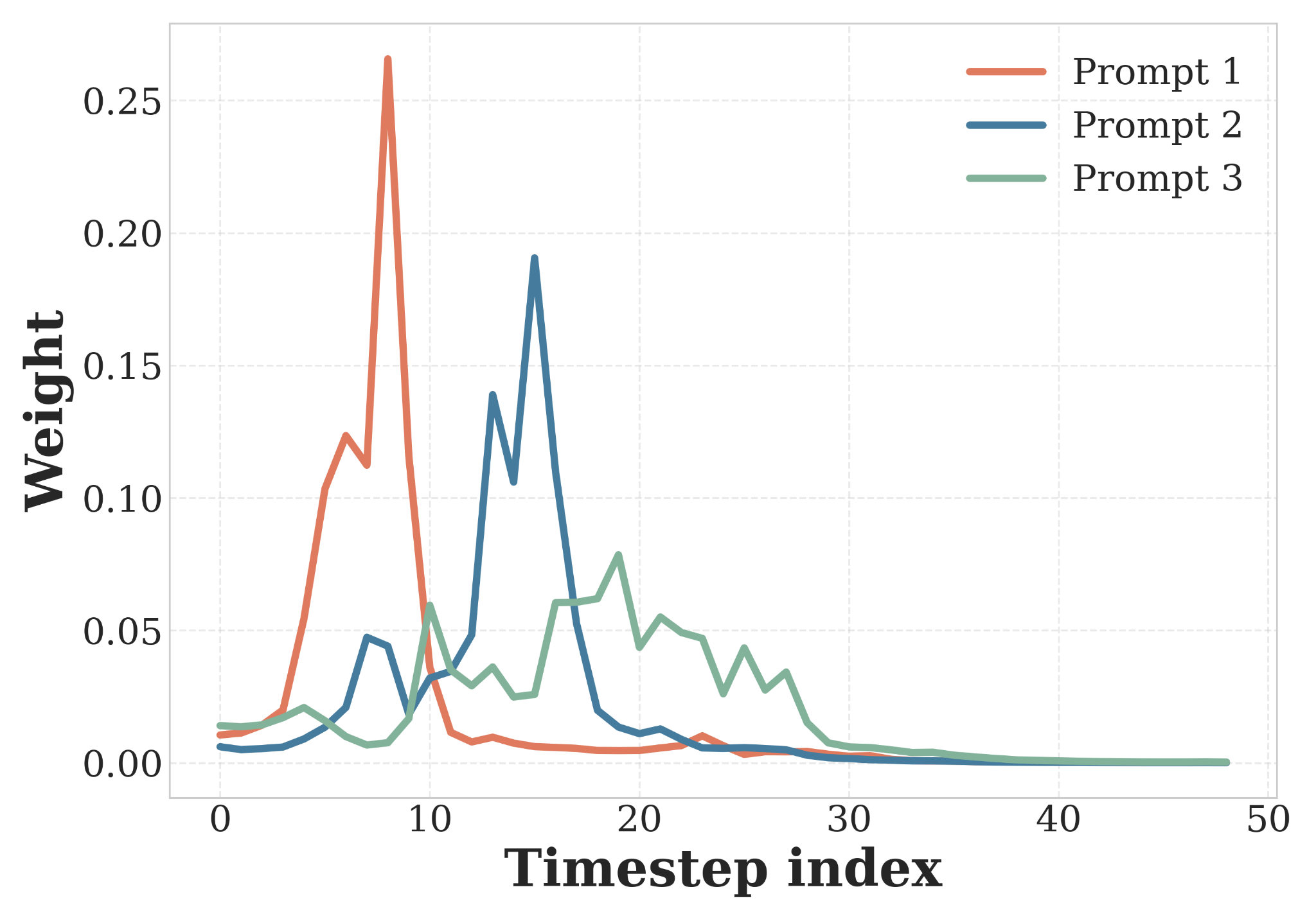}
        \caption*{(b)}
        \label{fig:prompt_weight}
    \end{minipage}
    \vspace{-0.8em}
    \caption{
    \textbf{Heuristic weighting and prompt-wise latent progress.}
    (a) Power-based early emphasis redistributes rewards according to exponent \(p\).
    (b) Prompt-wise contribution patterns vary across trajectories estimated by CoCA, illustrating the failure mode of fixed temporal priors and motivating adaptive credit assignment.
    }
    \label{fig:power_prompt}
\end{figure}

\subsection{Contribution-based Credit Assignment}
\label{ssec:coca}

To address the limitations of sparse rewards in diffusion-based text-to-image generation, we introduce Contribution-based Credit Assignment (CoCA). It estimates the contribution of each denoising step and then redistributes the final reward accordingly, providing informative step-level signals for policy optimization while preserving the stationary points of the original objective under policy gradient optimization.  We first characterize when an intermediate denoising step can \emph{in principle} influence the terminal reward.

\begin{proposition}[Latent-level upper bound on potential reward influence]
\label{prop:latent-upperbound}
Fix a prompt $c$.
Assume the terminal reward $r(x_0,c)$ is locally $L$-Lipschitz in $x_0$, and the remaining denoising dynamics from an intermediate latent $x_{T-t}$ to the terminal latent are locally $K_t$-Lipschitz, i.e., $x_0 = g_t(x_{T-t})$ for a locally $K_t$-Lipschitz map $g_t$.
Then for any $x_{T-t}, x'_{T-t}$ in a neighborhood of the realized trajectory,
\begin{equation}
\big| r(g_t(x_{T-t}), c) - r(g_t(x'_{T-t}), c) \big|
\;\le\;
L K_t \, \| x_{T-t} - x'_{T-t} \|.
\end{equation}
\end{proposition}

This bound characterizes when an intermediate denoising step can potentially influence the terminal reward: only steps inducing sufficiently large latent deviations can have a non-negligible effect. However, Eq.~(10) provides an upper bound on potential influence rather than a criterion for reward attribution, as it does not distinguish reward-relevant directions from orthogonal variations.

In CoCA, we therefore adopt a conservative proxy for credit redistribution referenced to the realized terminal outcome. Since the reward is defined solely on the final sample, intermediate steps can be related to it only through their consistency with the terminal latent. While the latent change magnitude $\|x_{T-t} - x_{T-(t-1)}\|$ reflects step activity, it is outcome-agnostic and conflates reward-relevant progress with orthogonal fluctuations.

Accordingly, we measure similarity progression with respect to the realized terminal latent $x_0$, emphasizing trajectory components aligned with the final outcome. This yields a trajectory-conditional explanatory notion of step importance: conditioned on a realized trajectory, steps that increase alignment with the terminal latent receive higher relative credit. Importantly, this is not a causal or counterfactual attribution, but a practical weighting signal for redistributing the terminal reward under policy optimization.
\begin{equation}
\label{eq:sim}
\begin{aligned}
\Delta \mathrm{Sim}_t
&= \mathrm{Sim}_t - \mathrm{Sim}_{t-1}, \\[2pt]
\mathrm{Sim}_t
&= \frac{\langle x_{T-t}, x_0 \rangle}
{\|x_{T-t}\| \cdot \|x_0\|}.
\end{aligned}
\end{equation}

To reduce fluctuations in per-step cosine similarity, we propose a \textbf{fixed window smoothing strategy}, segmenting $T$ timesteps into non-overlapping windows of fixed size $W$. Let the $i$-th window covers timesteps $t \in \{t_i, t_i + 1, \dots, t_i + W - 1\}$, where $t_i = i \cdot W + 1$ denotes the starting timestep of the $i$-th window. The average cosine similarity within the $i$-th window is defined as:

\begin{equation}
\begin{aligned}
\bar{\mathrm{Sim}}_i
&= \frac{1}{W} \sum_{t=t_i}^{t_i + W - 1} \mathrm{Sim}_t, \\[2pt]
&\qquad i \in \{0, 1, \dots, \lfloor T / W \rfloor\}.
\end{aligned}
\end{equation}

The smoothed progress score of window $i$ is then defined as $\Delta \bar{Sim}_i = \bar{Sim}_i - \bar{Sim}_{i-1}, \quad \text{where} \quad i \in \{1, \dots, \lfloor T / W \rfloor\}$. For the first window, we compute $\Delta \bar{Sim}_0 = \bar{Sim}_{0} - Sim_{0}$. This window-based smoothing yields a more stable estimate of timestep-wise latent progress during the denoising process.

\paragraph{Step-level Reward Redistribution.}
We redistribute the final reward $r(x_0, c)$ to each denoising step based on the estimated latent-progress proxy. Specifically, we normalize the progress scores $\Delta \bar{Sim}_i$ obtained from fixed window smoothing to compute the weight $w_t$ for each timestep $t$ within the $i$-th window. We further visualize the resulting CoCA weighting schedule without window smoothing to illustrate how reward is allocated across timesteps in Figure~\ref{fig:power_prompt} (b). The step-wise reward $\hat{R}(s_t, a_t)$ is then computed as:
\begin{equation}
\begin{aligned}
\hat{R}(s_t, a_t) &= w_t \cdot r(x_0, c), \\[2pt]
\text{where}\quad
w_t &= \frac{\Delta \bar{Sim}_i}
{W \sum_{k=1}^{\lfloor T / W \rfloor} \Delta \bar{Sim}_k}.
\end{aligned}
\end{equation}

\paragraph{Reward Normalization.}

To stabilize training and reduce variance of step-level rewards, we adopt a reward normalization strategy both before and after reward redistribution. To preserve rankings and reduce reward variance across prompts before reward shaping, we apply \textbf{per-prompt normalization}. For each prompt $p$, we collect $G$ trajectory-level rewards \( \{ r^1, r^2, \dots, r^G \} \) from the old policy \( \pi_{\theta_{\text{old}}} \), and normalize each as $\hat{A}^g = \frac{r^g - \mu_p}{\sigma_p + \epsilon}, \quad \text{where } \mu_p = \mathrm{mean}(r),\ \sigma_p = \mathrm{std}(r),\ \epsilon=1e-6.$

To further capture temporal variations after reward shaping, we adopt \textbf{per-prompt per-timetep normalization}. Given timestep-wise rewards $\mathbf{r}^g = [r^g_0, r^g_1, \dots, r^g_{T-1}]$, we compute the average and standard deviation per sample: $\mu^g = \mathrm{mean}(r^g), \quad \sigma^g = \mathrm{std}(r^g).$ Then, we estimate prompt-level statistics:$\mu_p = \mathbb{E}_{g}[\mu^g], \quad \sigma_p = \sqrt{\mathbb{E}_{g}[(\mu^g)^2 + (\sigma^g)^2] - \mu_p^2}.$ Finally, the normalized reward at each timestep \( t \) is:$\hat{A}^g_t = \frac{r^g_t - \mu_p}{\sigma_p + \epsilon}.$ This approach stabilizes reward scales both across prompts and over time, enabling more robust policy updates.

\paragraph{Policy Gradient of the Step-level Reward.} We derive gradients of the objective after contribution-based credit assignment similar to Equation~\ref{eq:policy}. Intuitively, CoCA reweights the policy gradient at each step by the cumulative future contribution mass, emphasizing steps that align more strongly with the realized outcome. The policy gradient with our contribution-based credit assignment $\nabla_\theta\mathcal{J}_{\text{CoCA}}(\pi_\theta)$ can be expressed as ( $\mathbb{E} \equiv \mathbb{E}_{\tau \sim \pi_\theta}$.):
    
\begin{equation}
\label{eq:coca-gradient}
\begin{aligned}
    &\mathbb{E}\left[\left(\sum_{t'=0}^{T-1}\hat{R}(s_{t'}, a_{t'})\right)\sum_{t=0}^{T-1} 
\nabla_\theta\log p_\theta(x_{T-t-1}|x_{T-t},c)\right]
\\&=\mathbb{E}\left[\sum_{t=0}^{T-1} \left(\sum_{t'=t}^{T-1}w_{t'}\right)r(x_0,c)\nabla_\theta\log p_\theta(x_{T-t-1}|x_{T-t},c)\right]    
 \end{aligned}
\end{equation}

\paragraph{Policy-Gradient Consistency under Reward Redistribution.}
We note an important optimization property of CoCA: redistributing the terminal reward across steps does not bias the expected policy gradient.

Specifically, if the redistributed step rewards satisfy
\begin{equation}
\sum_{t=0}^{T-1} \hat{R}(s_t,a_t) = r(x_0,c)
\label{eq:reward-conservation}
\end{equation}
for every trajectory, then the two objectives are identical:
\begin{equation}
\mathbb{E}_{\tau \sim \pi_\theta}[r(x_0,c)]
=
\mathbb{E}_{\tau \sim \pi_\theta}\!\left[\sum_{t=0}^{T-1} \hat{R}(s_t,a_t)\right].
\end{equation}
As a result, they share the same policy-gradient stationary points, and optimizing the redistributed objective preserves the original policy gradient.

In our implementation, the normalization used in reward redistribution ensures
$\sum_{t=0}^{T-1} w_t = 1$, so that~\eqref{eq:reward-conservation} holds.

%% file: sec/5_experiments.tex
\begin{figure*}
    \centering
    \includegraphics[width=0.95\linewidth]{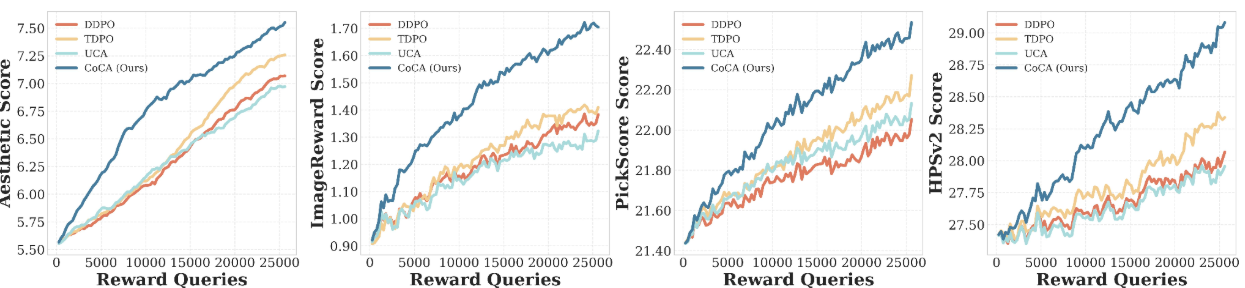}
    \caption{\textbf{Comparison of sample efficiency on the REINFORCE-based setting.} Reward functions (From left to right: (a) Aesthetic Score, (b) ImageReward Score, (c) PickScore, (d) HPSv2 Score) are evaluated to compare DDPO, TDPO, UCA, and our method. CoCA exhibits consistently steeper learning curves than both trajectory-level and step-level baselines, and shows faster smaple efficiency.
    }
    \label{fig:reward-curve}
\end{figure*}

\begin{figure*}[t]
  \centering
  \includegraphics[width=0.8\textwidth]{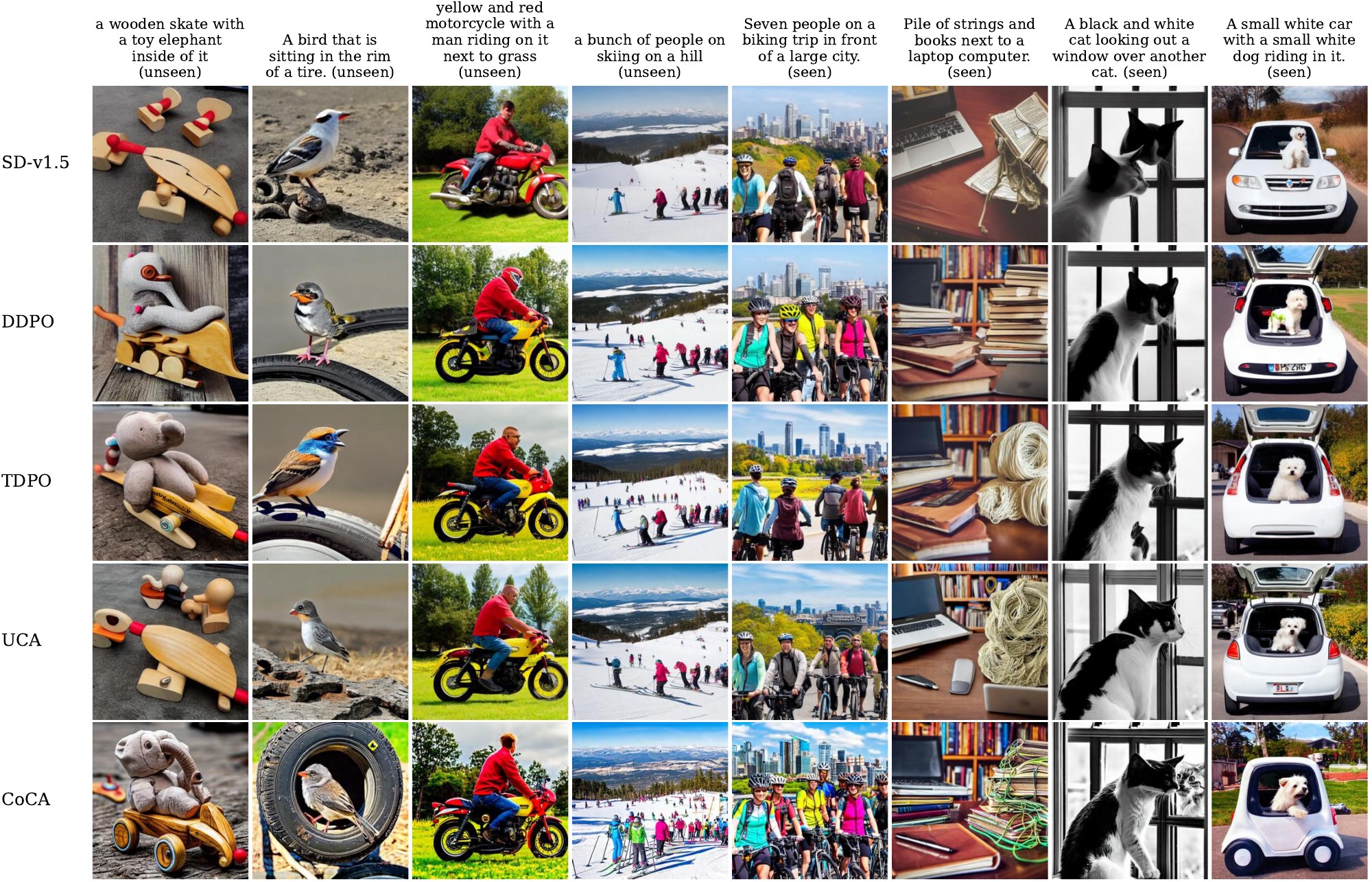}
  \caption{\textbf{Qualitative comparison on seen and unseen HPSv2 prompts.} CoCA achieves stronger semantic alignment and compositional accuracy across both prompt types.}
  \label{fig:quan-hpsv2}
  \vspace{-1em}
\end{figure*}

\begin{table*}[t]
\centering
\small
\caption{\textbf{Quantitative results on the Reinfocement-based setting.}
HPSv2 and PickScore are multiplied by 100 for readability, while ImageReward remains in its original scale.}
\begin{tabular}{lcccccccc}
\toprule
\multirow{2}{*}{\textbf{Method}} 
& \multicolumn{2}{c}{\textbf{Aesthetic ($\uparrow$)}} 
& \multicolumn{2}{c}{\textbf{PickScore ($\uparrow$)}} 
& \multicolumn{2}{c}{\textbf{ImageReward ($\uparrow$)}} 
& \multicolumn{2}{c}{\textbf{HPSv2 ($\uparrow$)}} \\
\cmidrule(lr){2-3} \cmidrule(lr){4-5} \cmidrule(lr){6-7} \cmidrule(lr){8-9}
 & \textbf{Train} & \textbf{Eval} & \textbf{Train} & \textbf{Eval} & \textbf{Train} & \textbf{Eval} & \textbf{Train} & \textbf{Eval} \\
\midrule
SD-v1.5~\citep{sd2022} & 5.57 & 5.63 & 21.46 & 21.67 & 0.92 & 0.53 & 27.37 & 27.69 \\
DDPO~\citep{ddpo2023} & 7.11 & 6.83 & 21.95 & 21.81 & 1.38 & 1.01 & 27.95 & 28.44 \\
UCA & 6.97 & 6.77 & 22.04 & 21.85 & 1.25 & 0.89 & 28.22 & 28.37 \\
TDPO~\citep{tdpo2024} & 7.30 & 7.27 & 22.21 & 21.96 & 1.44 & 0.94 & 28.30 & 28.61 \\
\textbf{CoCA (Ours)} &
\textbf{7.58} & 
\textbf{7.41} &
\textbf{22.49} & 
\textbf{22.27} &
\textbf{1.69} & 
\textbf{1.32} &
\textbf{29.08} & 
\textbf{29.15} \\
\bottomrule
\end{tabular}
\vspace{-1em}
\label{tab:reinf-result}
\end{table*}

\begin{table*}[t]
\centering
\small
\setlength{\tabcolsep}{6pt}
\caption{\textbf{Quantitative results on the GRPO-based setting.}
All metric values except Aesthetic are multiplied by 100 for readability.
\textcolor{red}{Red} numbers indicate absolute improvements of CoCA over DanceGRPO.}
\label{tab:grpo}
\begin{tabular}{lcccc}
\toprule
\textbf{Method} & \textbf{HPSv2 ($\uparrow$)} & \textbf{CLIP Score ($\uparrow$)} & \textbf{PickScore ($\uparrow$)} & \textbf{Aesthetic ($\uparrow$)} \\
\midrule
DanceGRPO~\cite{dancegrpo2025} (HPSv2) & 37.13 & 35.89 & 22.32 & 6.37 \\
\textbf{CoCA (HPSv2)} &
\textbf{37.20} {\color{red}{(+0.07)}} &
\textbf{36.66} {\color{red}{(+0.77)}} &
\textbf{22.47} {\color{red}{(+0.15)}} &
6.34 \\
\midrule
DanceGRPO~\cite{dancegrpo2025} (HPSv2 + CLIP) & 32.47 & 39.70 & 21.98 & 6.13 \\
\textbf{CoCA (HPSv2 + CLIP)} &
\textbf{33.84} {\color{red}{(+1.37)}} &
\textbf{40.96} {\color{red}{(+1.26)}} &
\textbf{22.40} {\color{red}{(+0.42)}} &
\textbf{6.21} {\color{red}{(+0.08)}} \\
\bottomrule
\end{tabular}
\end{table*}

\section{Experiments}
\label{sec:exp}
In this section, we present comprehensive experiments demonstrating that our method serves as a general credit assignment framework applicable to both the REINFORCE-based~\citep{mcpg2004} policy optimization method DDPO~\citep{ddpo2023} and the GRPO-based method DanceGRPO~\citep{dancegrpo2025}.
Experimental results show that leveraging step-level rewards without any external models substantially improves sampling efficiency and enhances generalization to both unseen reward functions and unseen prompts.

\subsection{Implementation Details}
\label{ssec:imp}

\paragraph{REINFORCE-based setting.}
Here we compare CoCA with representative trajectory-level and step-level approaches based on DDPO~\citep{ddpo2023}: (1) SD-v1.5~\cite{sd2022}: Pretrained backbone before RL fine-tuning. (2) DDPO~\cite{ddpo2023}: the commonly used trajectory-level reward optimization algorithm, illustrated in~\ref{fig:intro-main}. (3) TDPO~\cite{tdpo2024}: A state-of-the-art step-level variant that learns a critic to estimate per-step baselines. (4) UCA: A uniform credit assignment baseline that converts the terminal reward into a simple step-level signal, assigning $\hat{R}(s_t, a_t) = r(x_0, c)/T$ to each timestep. As such, it serves as a naive step-level discounting scheme rather than modeling trajectory-dependent credit.

For the REINFORCE-based setting, Models are fine-tuned separately on four commonly used image-quality rewards: Aesthetic~\citep{aesthetic2022}, PickScore~\citep{pick2023}, ImageReward~\citep{imagereward2023}, and HPSv2~\citep{hps2023}.
For Aesthetic, PickScore, and ImageReward, training is conducted on 45 animal categories, with evaluation on 8 unseen categories to assess cross-category generalization.
For HPSv2, we follow its original Dataset protocol, training on 750 prompts and evaluating on 50 held-out prompts.
\vspace{-1em}
\paragraph{GRPO-based setting.}
We further evaluate CoCA on top of the stronger GRPO~\citep{grpo} optimization framework: DanceGRPO~\citep{dancegrpo2025}: A GRPO-based method supporting both single- and multi-reward optimization. The detailed experimental settings are provided in Appendix~\ref{sec:apdx-adimp}).

For a stronger GRPO-based evaluation, we adopt the official benchmark of HPSv2~\citep{hps2023}. The training split contains 103,700 prompts, while the test split contains 400 prompts.
Following prior work~\citep{dancegrpo2025}, we benchmark under both single-reward (HPSv2) and multi-reward (HPSv2 + CLIP~\citep{clip2021}) configurations.

\subsection{
Trajectory-level vs. 
 Step-level Rewards}
\label{ssec:comp}
\paragraph{Sample Efficiency in Reward Optimization.}

We assess the performance of reward-based finetuning algorithms for diffusion models using sample efficiency, where efficiency is measured by the improvement in generation quality per reward query (a single reward evaluation of a generated image). Figure~\ref{fig:reward-curve} shows the learning curves under the DDPO setting across four reward functions, plotting reward against the number of queries.
Across all metrics, CoCA exhibits consistently steeper learning curves than both trajectory-level and step-level baselines, achieving 25\%–100\% faster convergence on average.
This demonstrates that CoCA learns more effectively from limited feedback and delivers substantially higher sample efficiency.

\paragraph{Quantitative and Qualitative Results.}
In the REINFORCE-based setting (Table~\ref{tab:reinf-result}), CoCA consistently achieves the best performance across nearly all reward functions, outperforming both trajectory-level (DDPO) and step-level (TDPO, UCA) baselines. The qualitative results shown in Figure~\ref{fig:quan-hpsv2} demonstrate that CoCA produces more faithful attributes and stronger compositional consistency across both seen and unseen prompts.
In the GRPO-based setting (Table~\ref{tab:grpo}), CoCA likewise delivers the strongest results, particularly under multi-reward optimization. CoCA also generalizes the improvement to reward metrics it was not trained on (such as PickScore), indicating that its gains stem from genuine image-quality improvements rather than reward hacking.
This advantage arises from CoCA’s contribution-based credit assignment, which offers stable, informative step-level signals that integrate multiple rewards more effectively.

\subsection{Ablation Study}
\label{ssec:ab}
\begin{figure}[t]
    \centering
    \includegraphics[width=0.55\linewidth]{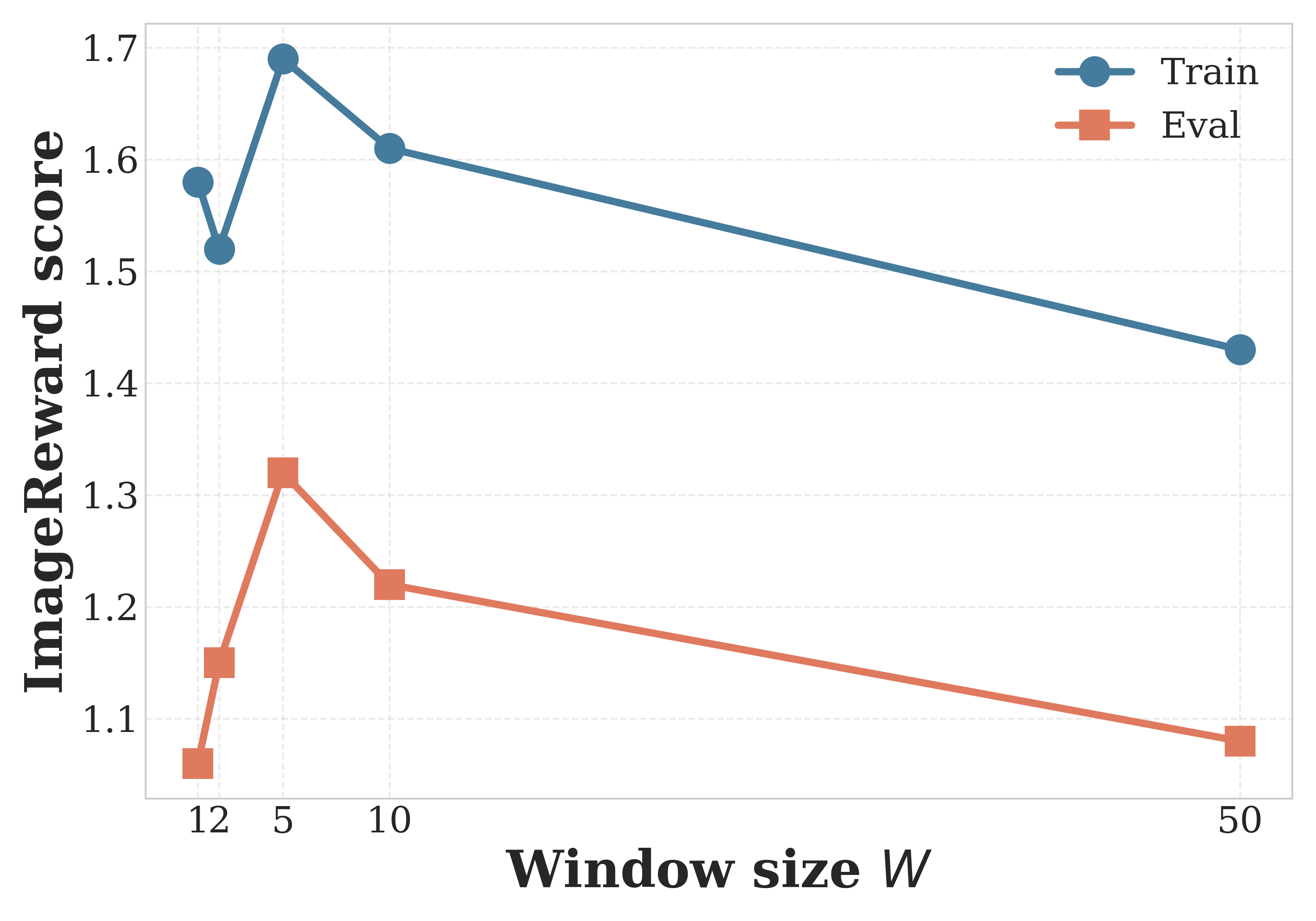}
    \caption{\textbf{Ablation on window size.} A moderate window size ($W{=}5$) yields the best performance.}
    \label{fig:ab-window}
    \vspace{-15pt}
\end{figure}

\paragraph{Ablation on Fixed-Window Smoothing.}
We investigate the impact of the smoothing window by different window sizes $W \in \{1, 2, 5, 10, 50\}$ under the ImageReward objective. As shown in Fig.~\ref{fig:ab-window} and Table~\ref{tab:ab-study}(a), a moderate window size of $W{=}5$ consistently provides the best performance. Therefore, we adopt $W{=}5$ for all subsequent experiments.

\paragraph{Ablation on Distance Metrics.} 
To estimate each step’s contribution, we measure the distance between intermediate and final latent embeddings using either cosine similarity or $\ell_2$ distance. As shown in Table~\ref{tab:ab-study}~(a), cosine similarity yields slightly better performance, so we adopt it as our default metric. We further compare these two metrics against an alternative approach that uses the reward of the predicted $\hat{x}_0$ at timestep $t$. However, relying on intermediate-step rewards performs significantly worse: it increases reward-query overhead and yields unstable estimates, since reward models struggle to assess noisy intermediate predictions.

\paragraph{Ablation on Reward Normalization.}
We analyze the effect of two normalization strategies: (1) per-prompt normalization before redistribution (P), and (2) per-prompt per-timestep normalization after redistribution (PT). 
As shown in Table~\ref{tab:ab-study}(b), each normalization individually improves stability, while combining both achieves the best performance.

\begin{table}[t]
\centering
\caption{
\textbf{Ablation study.} 
(a) Ablation study on distance metrics. 
(b) Ablation study on reward normalization.}
\label{tab:ab-study}
\begin{minipage}[t]{0.46\linewidth}
\centering(a)
\begin{tabular}{lcc}
    \toprule
    Sim & Train & Eval \\
    \midrule
    reward & 6.68 & 6.56 \\
    $\ell_2$ & 7.53 & 7.36 \\
    \textbf{cosine} & \textbf{7.58} & \textbf{7.41} \\
    \bottomrule
    \label{ab:sim}
\end{tabular}
\end{minipage}
\hfill
\begin{minipage}[t]{0.46\linewidth}
\centering(b)
\vspace{1em}
\begin{tabular}{llcc}
    \toprule
    P & PT & Train & Eval \\
    \midrule
    \checkmark &  & 7.50 & 7.29 \\
     & \checkmark & 4.11 & 3.80 \\
    \checkmark & \checkmark & \textbf{7.58} & \textbf{7.41} \\
    \bottomrule
    \label{ab:reg}
\end{tabular}
\end{minipage}
\vspace{-30pt}
\end{table}

%% file: sec/6_conclusion.tex
\section{Conclusion}
\label{sec:conclusion}
In this work, we observe that a credit assignment mismatch caused by sparse trajectory-level rewards exists in RL-based diffusion model fine-tuning methods. To solve this, we propose CoCA, a contribution-based credit assignment framework that estimates the impact of each step on the final image and redistributes rewards accordingly. Without introducing extra networks or heuristics, CoCA significantly improves sample efficiency, convergence speed, and generalization across multiple human preference reward functions. 

%% file: sec/7_suppl.tex
\newpage
\appendix
\onecolumn
\appendix

\section{Derivations}
\label{sec:apdx-derivation}
\begin{proof}
We aim to compute the gradient of the expected cumulative reward under the Contribution-based credit assignment (CoCA) setting. The objective is defined as:
\begin{equation}
\begin{aligned}
    \nabla_\theta\mathcal{J}_{\text{CoCA}}(\pi_\theta)
    = \nabla_\theta \mathbb{E}_{\tau \sim \pi_\theta}\left[\sum_{t=0}^{T-1}\hat{R}(s_t, a_t)\right]
    = \sum_{t'=0}^{T-1} \nabla_\theta \mathbb{E}_{\tau^{t'}}\left[\hat{R}(s_{t'}, a_{t'})\right],
\end{aligned}
\end{equation}
where we decompose the expectation using the linearity of the gradient operator. Here, $\tau^{t'}$ denotes the partial trajectory up to time step $t'$, and $\hat{R}(s_{t'}, a_{t'})$ only depends on past decisions due to the Markov property of the environment.

By applying the policy gradient theorem~\citep{policygradient1999} at each time step:
\begin{equation}
\begin{aligned}
    \nabla_\theta \mathbb{E}_{\tau^{t'}}\bigg[\hat{R}(s_{t'}, a_{t'})\bigg]
    = \mathbb{E}_{\tau^{t'}}\left[\hat{R}(s_{t'}, a_{t'}) \sum_{t=0}^{t'} \nabla_\theta \log p_\theta(x_{T-t-1} \mid x_{T-t}, c)\right],
\end{aligned}
\end{equation}
where we express the trajectory distribution using the reverse-time transition probabilities of the diffusion model, i.e., $p_\theta(x_{T-t-1} \mid x_{T-t}, c)$.

Combining the terms across all time steps:
\begin{equation}
\begin{aligned}
\nabla_\theta \mathcal{J}_{\text{CoCA}}(\pi_\theta) &= \sum_{t'=0}^{T-1}
   \mathbb{E}_{\tau^{t'}} \!\left[
      \hat{R}(s_{t'}, a_{t'})
      \sum_{t=0}^{t'}
      \nabla_\theta \log p_\theta(x_{T-t-1} \mid x_{T-t}, c)
   \right] \\
&= \mathbb{E}_{\tau \sim \pi_\theta} \!\left[
      \sum_{t'=0}^{T-1} \hat{R}(s_{t'}, a_{t'}) 
   \right. \left.
      \sum_{t=0}^{t'}
      \nabla_\theta \log p_\theta(x_{T-t-1} \mid x_{T-t}, c)
   \right].
\end{aligned}
\end{equation}

We now rearrange the summation over $(t, t')$ by swapping the order:
\begin{equation}
\begin{aligned}
    \nabla_\theta \mathcal{J}_{\text{CoCA}}(\pi_\theta) \\&=
\mathbb{E}_{\tau \sim \pi_\theta}
\left[
\sum_{t=0}^{T-1}
\left(
\sum_{t'=t}^{T-1}
\hat{R}(s_{t'},a_{t'})
\right)
\nabla_\theta
\log p_\theta(x_{T-t-1}\mid x_{T-t},c)
\right].
\end{aligned}
\end{equation}

This rearrangement can be understood via the following illustrative derivation:
\begin{equation}
\begin{aligned}
&r_0(f_0) + r_1(f_0 + f_1) + \dots + r_{T-1}(f_0 + \dots + f_{T-1}) \\
&= (r_0 + \dots + r_{T-1}) f_0 + (r_1 + \dots + r_{T-1}) f_1 \quad+ \dots + r_{T-1} f_{T-1},
\end{aligned}
\end{equation}
where $r_t \coloneqq \hat{R}(s_t, a_t)$ and $f_t \coloneqq \nabla_\theta \log p_\theta(x_{T-t-1} \mid x_{T-t}, c)$.

Finally, under the CoCA assumption that each reward at time $t$ is assigned by its contribution to the final image, i.e., $\hat{R}(s_t, a_t) = w_t r(x_0, c)$, we obtain:
\begin{equation}
\begin{aligned}
\nabla_\theta \mathcal{J}_{\text{CoCA}}(\pi_\theta)
&= \mathbb{E}_{\tau \sim \pi_\theta} \bigg[\sum_{t=0}^{T-1}\Big(\sum_{t'=t}^{T-1} w_{t'}\Big)r(x_0, c) \nabla_\theta \log p_\theta(x_{T-t-1} \mid x_{T-t}, c) \bigg].
\end{aligned}
\end{equation}

which completes the proof.
\end{proof}

\label{sec:apdx-proof}

\section{Additional Implementaion Details}
\label{sec:apdx-adimp}
\subsection{Configuration and Reward Functions}
\label{ssec:apdx-config}

In our experiments, we adopt Stable Diffusion v1.5~\cite{sd2022} as the base generative model for all methods, ensuring a fair comparison across different reward functions and training algorithms. All models are fine-tuned using Low-Rank Adaptation (LoRA)~\cite{lora2022} applied to the attention layers in the UNet backbone~\cite{unet2015}, significantly reducing training overhead while maintaining model performance.
\vspace{-8pt}

\paragraph{DDPO~\cite{ddpo2023} and TDPO~\cite{tdpo2024} implementations.}  
We build upon the official PyTorch implementation of DDPO to reproduce both DDPO and TDPO results. Our experiments are run on a system with 4 NVIDIA RTX 4090 GPUs (24GB memory each), and we provide configurations specifically adapted for this hardware setting. To address memory constraints, we apply \textit{gradient checkpointing}~\cite{gradientchpt2016} to the critic model in TDPO, enabling larger batch sizes without exceeding GPU memory limits.
\vspace{-8pt}

\paragraph{UCA and CoCA implementations.}  
UCA and CoCA are implemented in the same training framework and are also trained on Stable Diffusion v1.5 using LoRA fine-tuning. All methods share the same sampling and training settings unless otherwise stated, ensuring consistency in comparison across different algorithms.
\vspace{-8pt}

\paragraph{Reward functions.}  
We evaluate all methods on four different reward functions: \textit{Aesthetic}~\cite{aesthetic2022}, \textit{PickScore}~\cite{pick2023}, \textit{ImageReward}~\cite{imagereward2023}, and \textit{HPSv2}~\cite{hps2023}. The hyperparameter settings for each reward are listed in Table~\ref{tab:reward-configs}. To align with the design of TDPO, we accelerate the gradient update frequency to $2 \times T$ (100) timesteps for all methods under the Aesthetic reward. Given the increased variance in returns caused by credit assignment, we further reduce the reward variants within a narrower range of $[-5 \times 10^{-5}, 5 \times 10^{-5}]$ to improve the stability of training.

\begin{table*}[htbp]
\centering
\caption{List of hyperparameter configurations for Aesthetic, PickScore, ImageReward, and HPSv2.}
\label{tab:reward-configs}
\begin{tabular}{l|c|c|c|c}
\toprule
\textbf{Hyperparameters} & \textbf{Aesthetic} & \textbf{PickScore} & \textbf{ImageReward} & \textbf{HPSv2} \\
\midrule
Random seed & 42 & 42 & 42 & 42 \\
Denoising timesteps ($T$) & 50 & 50 & 50 & 50 \\
Guidance scale & 5.0 & 5.0 & 5.0 & 5.0 \\
Policy learning rate & 1e-4 & 1e-4 & 1e-4 & 1e-4 \\
Policy clipping range & 5e-5 & 1e-4 & 1e-4 & 1e-4 \\
Maximum gradient norm & 1.0 & 1.0 & 1.0 & 1.0 \\
Optimizer & AdamW & AdamW & AdamW & AdamW \\
Optimizer weight decay & 1e-4 & 1e-4 & 1e-4 & 1e-4 \\
Optimizer $\beta_1$ & 0.9 & 0.9 & 0.9 & 0.9 \\
Optimizer $\beta_2$ & 0.999 & 0.999 & 0.999 & 0.999 \\
Optimizer $\epsilon$ & 1e-8 & 1e-8 & 1e-8 & 1e-8 \\
Sampling batch size & 16 & 16 & 16 & 16 \\
Samples per epoch & $256 $ & $256 $ & $256 $ & $256 $ \\
Training batch size & 4 & 4 & 4 & 4 \\
Gradient accumulation steps & 32 & 16 & 16 & 16 \\
Training steps per epoch & $128 $ & $64 $ & $64 $ & $64 $ \\
Gradient updates per epoch & $2 \times T$ & 4 & 4 & 4 \\
window size & 5 & 5 & 5 & 5 \\
\bottomrule
\end{tabular}
\end{table*}

\subsection{List of 8 Unseen Animals}
\label{ssec:apdx-unseen}

We evaluated the prompt generalization capabilities on 8 unseen animals following TDPO~\cite{tdpo2024}: snail, hippopotamus, cheetah, crocodile, lobster, octopus, elephant, and jellyfish. 

\section{Limitations and Future Work}
\label{sec:apdx-limit}

\paragraph{Limitation}
Although the CoCA algorithm demonstrates promising results in densifying rewards and accelerating the optimization process, it exhibits certain limitations. First, according to qualitative analysis, CoCA tends to induce rapid changes in the global structure of generated samples. While this behavior reflects the model's sensitivity to trajectory-level preferences, it may also lead to risks of over-saturation and over-sharpening, potentially harming sample naturalness. 

\paragraph{Future Work}
Future directions include a more in-depth investigation into the impact of different credit assignment strategies, particularly how they affect training dynamics and generation quality. In addition, integrating CoCA with Direct Preference Optimization (DPO) may offer a promising path toward more stable and interpretable preference learning. Another key direction lies in enhancing CoCA to incorporate explicit step-level preference signals, enabling more precise alignment between user feedback and step-wise optimization, and potentially bridging the gap between trajectory-level supervision and fine-grained control.

\section{Social Impacts}
\label{sec:apdx-impact}
This work contributes to improving text-to-image (T2I) generation in terms of both human preference alignment and instruction following. By introducing step-level reward by contribution-based credit assignment, our method allows T2I diffusion models to generate images that better align with nuanced human intentions, promoting more reliable and controllable human-AI interaction. This has potential applications in personalized content creation, assistive design, and other domains requiring fine-grained visual generation. Moreover,  we achieve competitive or superior performance with fewer training samples, leading to reduced energy consumption and improved training efficiency. This aligns with the broader goals of sustainable AI and responsible machine learning development.

\section{More Qualititive Results}
\label{sec:apdx-qua}
\subsection{More Qualitative Results on unseen animals}

\begin{figure*}
  \centering
  \includegraphics[width=\textwidth]{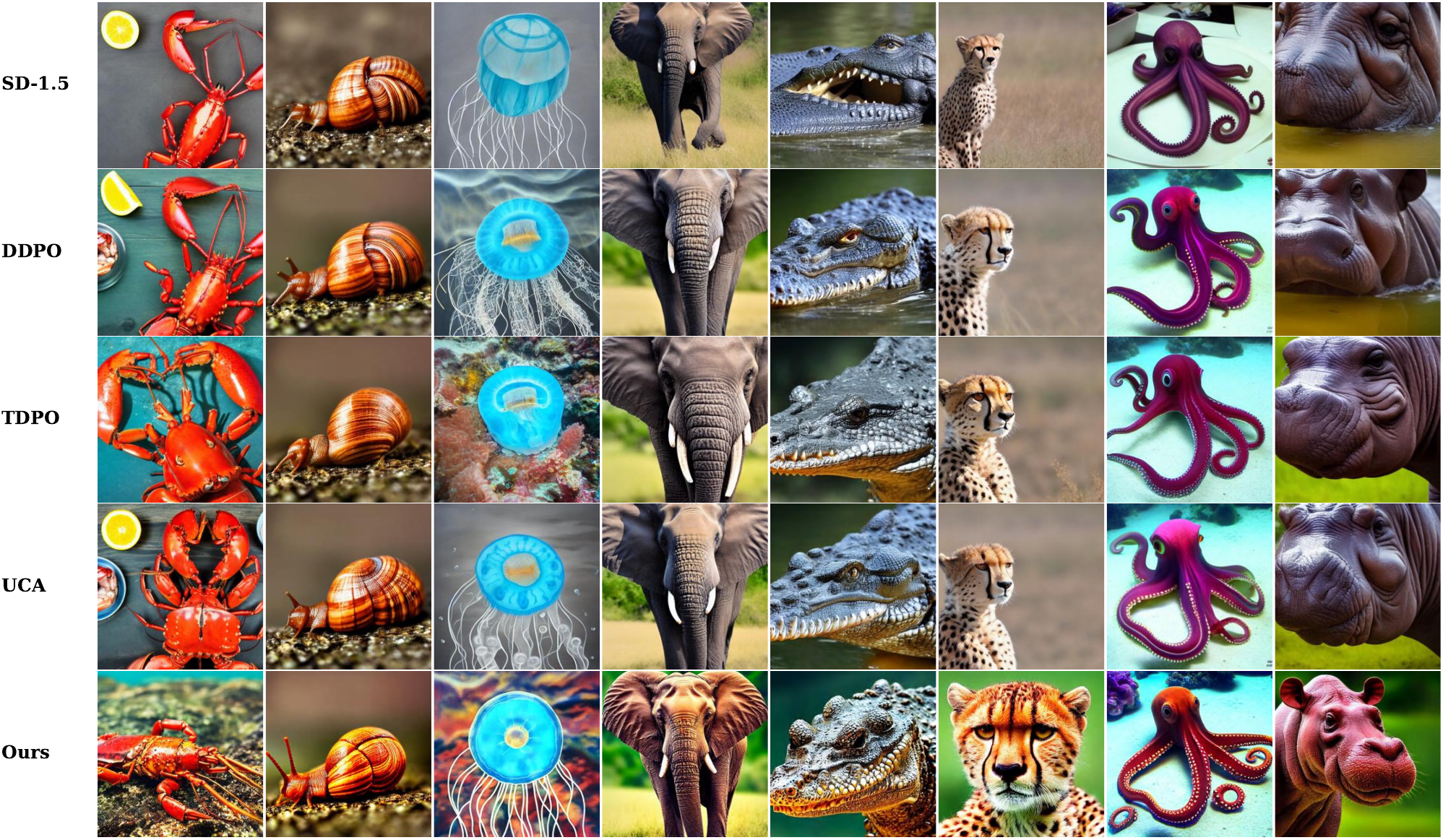}
  \caption{Qualitative comparison of unseen animals generated by SD-v1.5, DDPO, TDPO, UCA, CoCA trained on ImageReward.}
  \label{unseen-prompts}
\end{figure*}

We generate samples from baselines and our methods trained on the ImageReward reward function, as shown in Figure~\ref{unseen-prompts}. The results show that CoCA has better aesthetic quality and alignment with the given prompts.

\subsection{More Qualitative Results on HPSv2 and Qualitative Analysis}
\label{apdx:qua-analysis}
\begin{figure*}
  \centering
  \includegraphics[width=\textwidth]{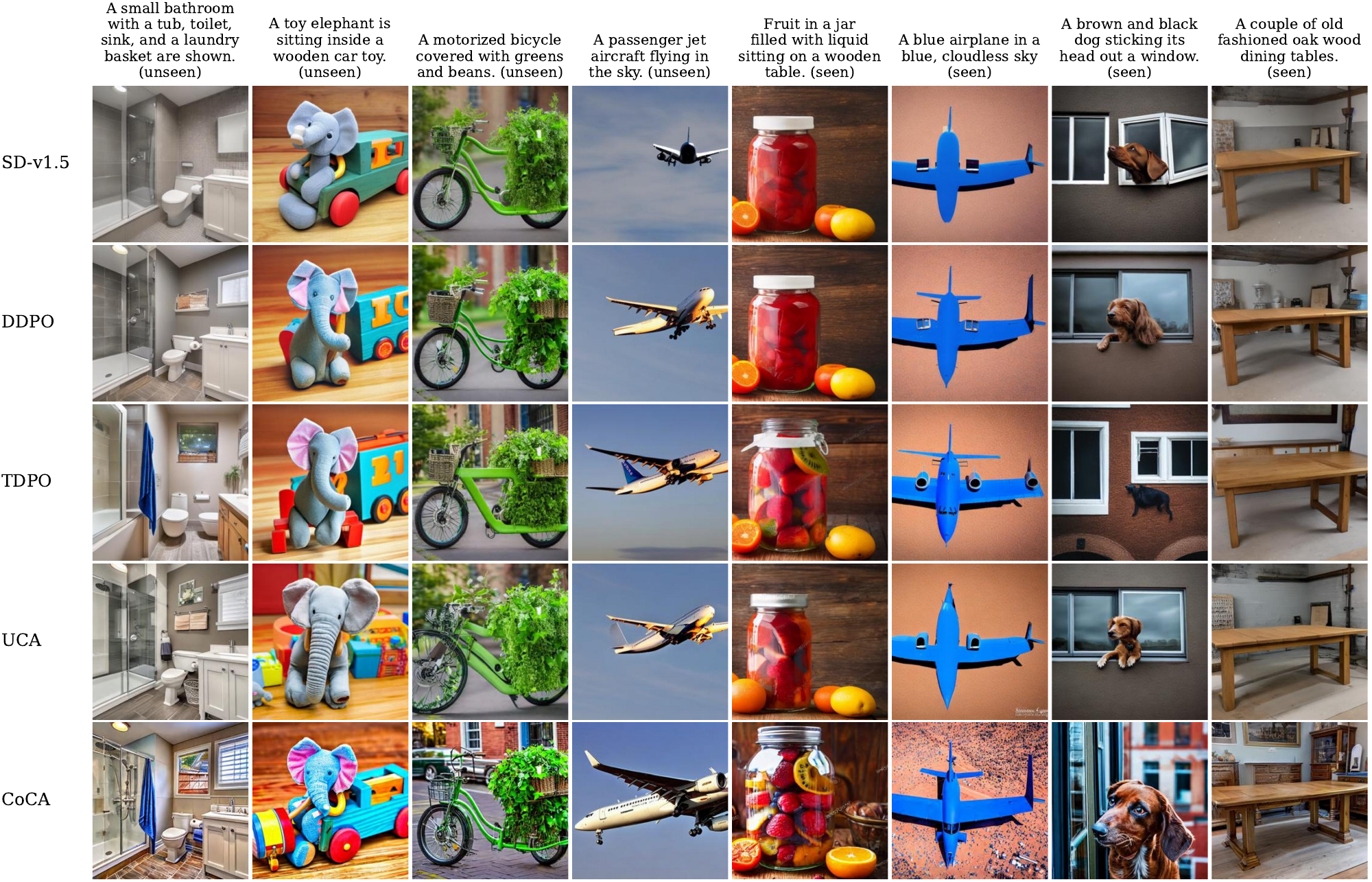}
  \caption{More qualitative comparison of unseen prompts from HPSv2 data generated by SD-v1.5, DDPO, TDPO, UCA, CoCA. }
  \label{unseen-hps}
\end{figure*}

We show more samples of all baselines, and CoCA generated fromthe  HPSv2 reward optimizing model in Figure~\ref{unseen-hps}. Here is a qualitative analysis for Figure~\ref{fig:quan-hpsv2} and Figure~\ref{unseen-hps}:

(1) Complex relationship: In Figure~\ref{fig:quan-hpsv2}, our method authentically generates the relationship "in" while others do not for the prompt "A bird that is sitting in the rim of a tire". For prompt "A black and white cat looking out a window over another cat, CoCA faithfully generates two cats looking at each other through the window, while other methods only generate a cat.

(2) Composition: In Figure~\ref{fig:quan-hpsv2}, CoCA accurately combines "a toy elephant" and "a wooden skate", while in Figure~\ref{unseen-hps}, CoCA also generates "a toy elephant" and "a wooden car toy" reasonably.

(3) Color: In Figure~\ref{fig:quan-hpsv2}, CoCA generates mixed color "yellow and red" without omission.
 
(4) Count: For prompt "Seven people on a biking trip in front of a large city." in Figure~\ref{fig:quan-hpsv2}, CoCA generates exact "seven" people, and others fail. 

\subsection{More Itermediate-step Samples from HPSv2}
\begin{figure*}
  \centering
  \includegraphics[width=0.9\textwidth]{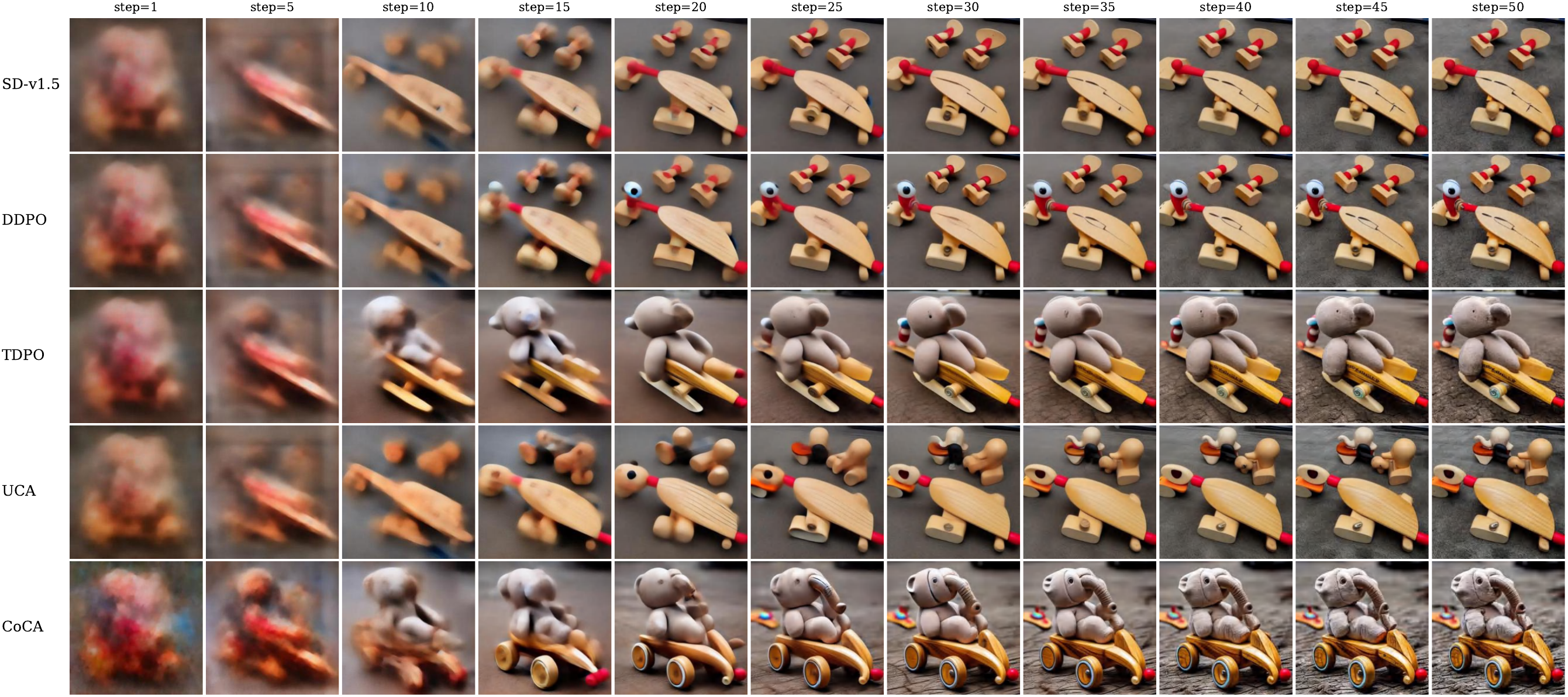}
  \caption{Qualitative comparison of samples of selected timesteps generated on prompt "a wooden skate with a toy elephant inside of it" by SD-v1.5, DDPO, TDPO, UCA, CoCA trained on HPSv2 reward function. Faster global structure changing is observed in samples from CoCA. }
  \label{step-result-48}

  \centering
  \includegraphics[width=0.9\textwidth]{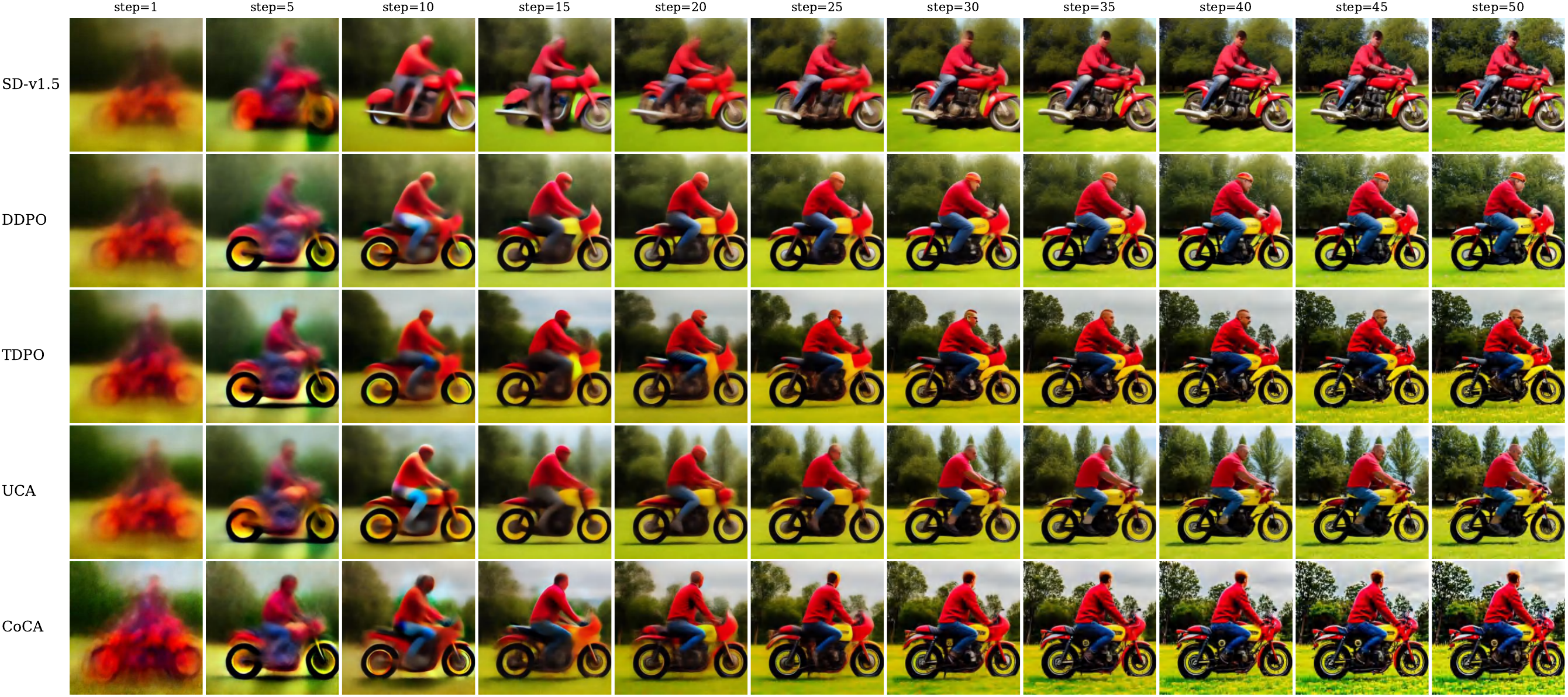}
  \caption{Qualitative comparison of samples generated on prompt "yellow and red motorcycle with a man riding on it next to grass" by SD-v1.5, DDPO, TDPO, UCA, CoCA trained on HPSv2 reward function.}
  \label{step-result-24}

  \centering
  \includegraphics[width=0.9\textwidth]{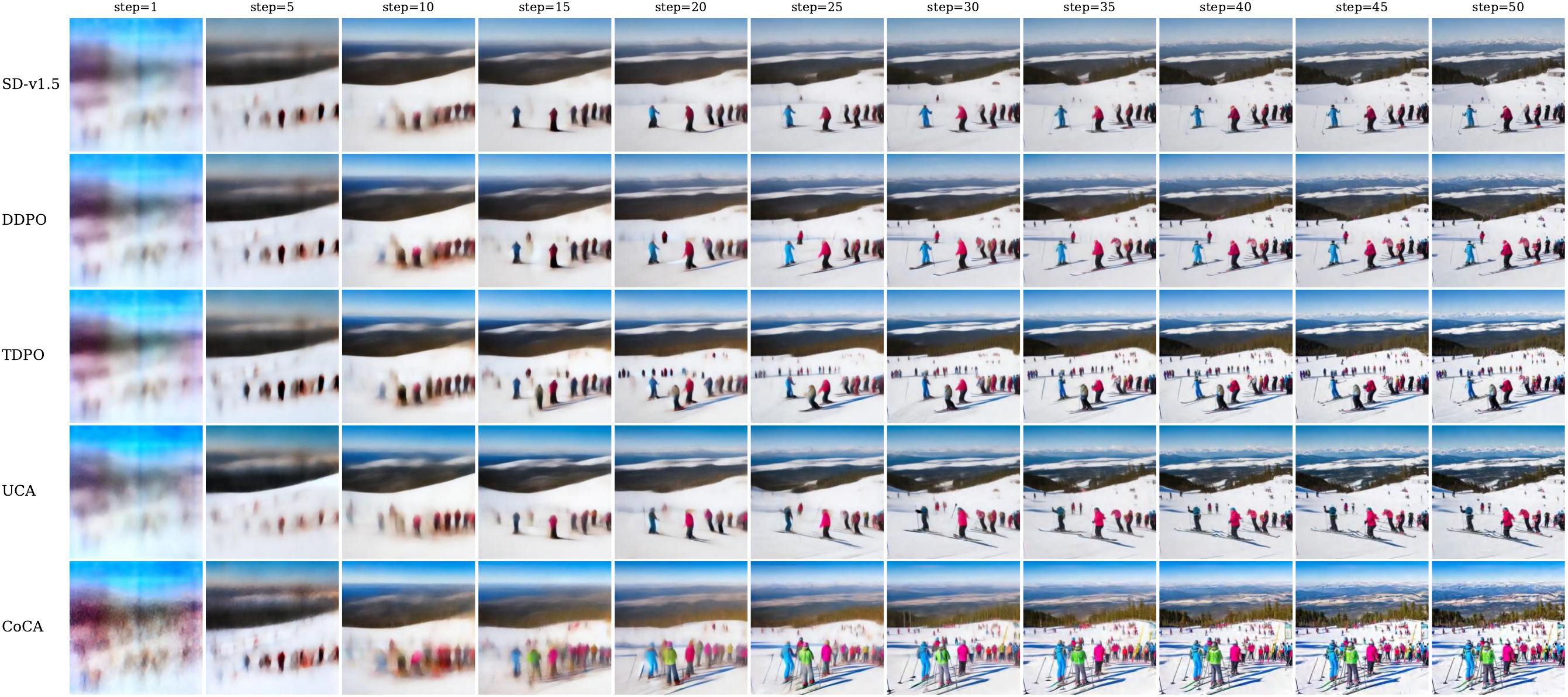}
  \caption{Qualitative comparison of samples generated on prompt "a bunch of people skiing on a hill" by SD-v1.5, DDPO, TDPO, UCA, CoCA trained on HPSv2 reward function.}
  \label{step-result-49}
\end{figure*}

\begin{figure*}[ht]
  \centering
  \includegraphics[width=0.9\textwidth]{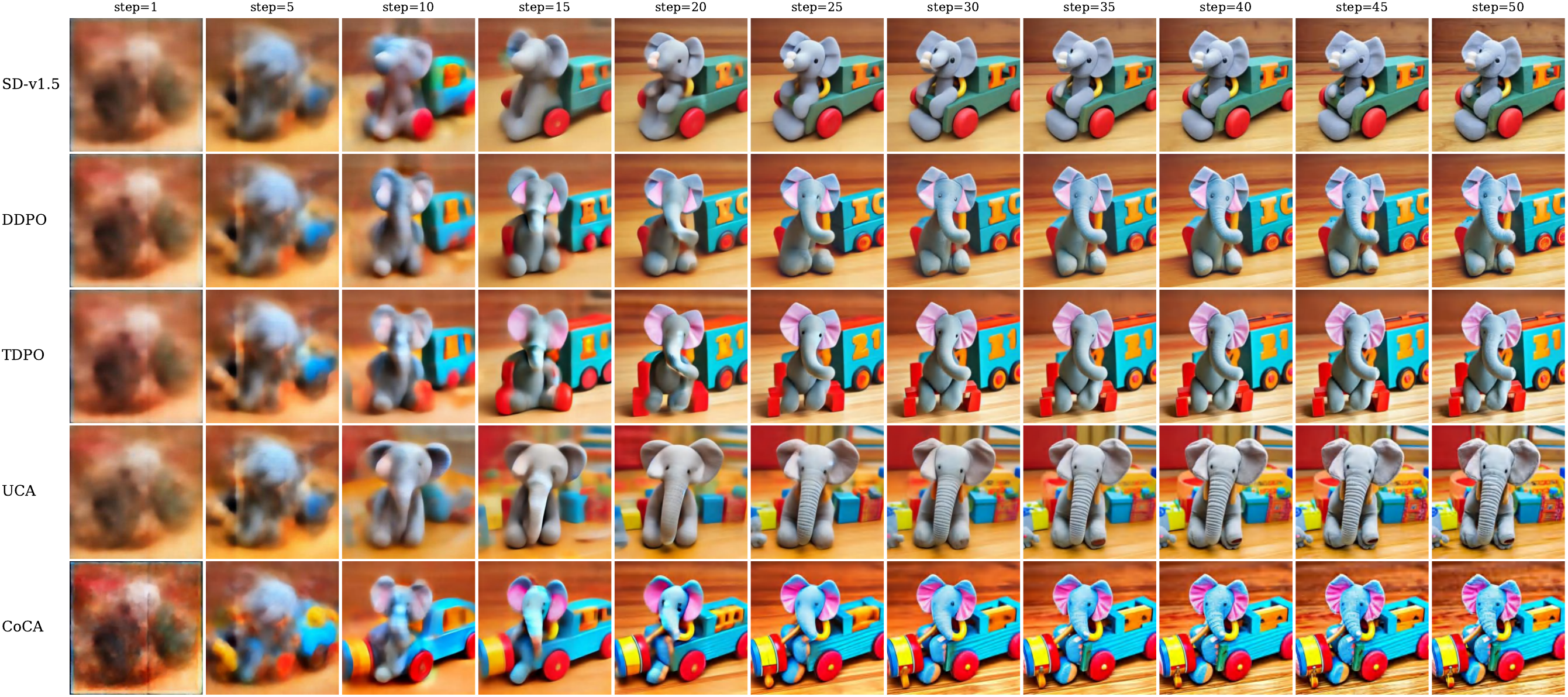}
  \caption{Qualitative comparison of samples generated on prompt "A toy elephant is sitting inside a wooden car toy." by SD-v1.5, DDPO, TDPO, UCA, CoCA trained on HPSv2 reward function.}
  \label{step-result-32}

  \centering
  \includegraphics[width=0.9\textwidth]{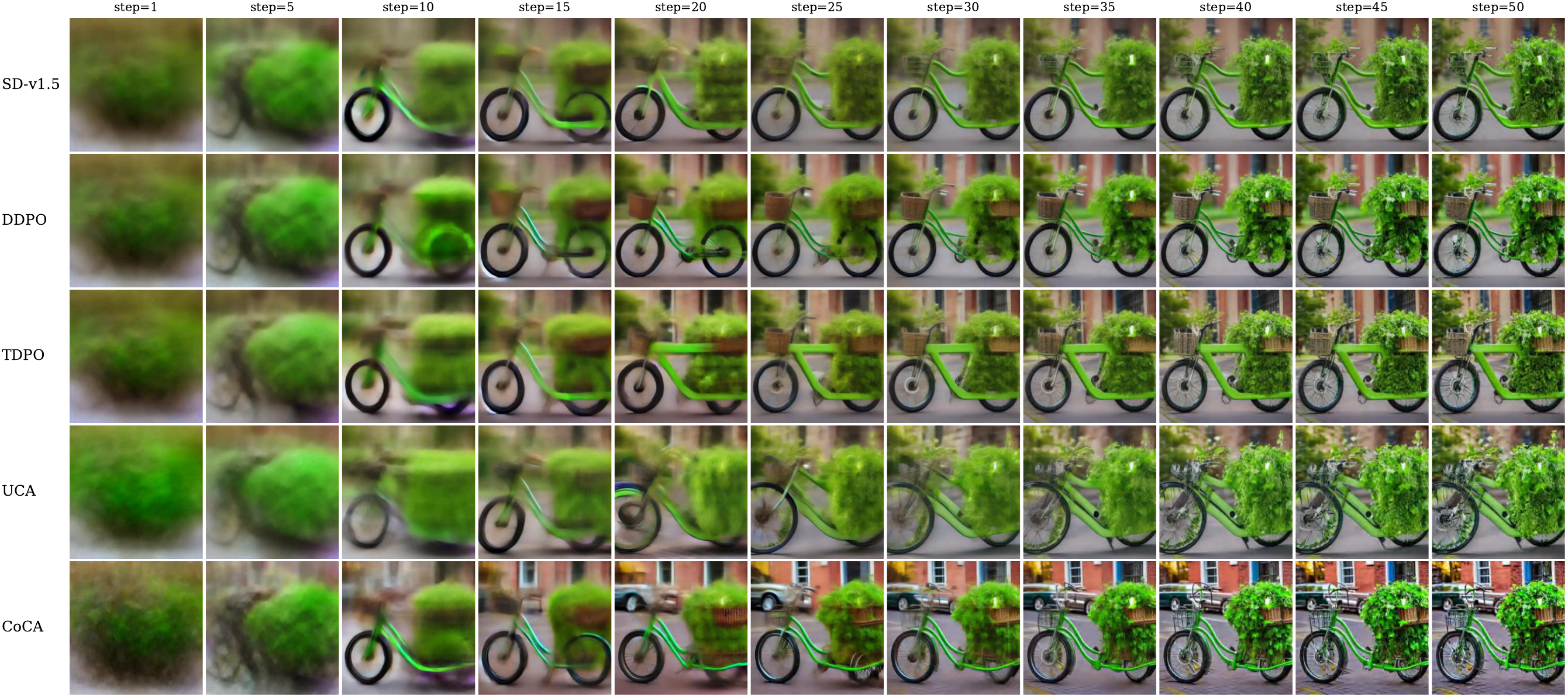}
  \caption{Qualitative comparison of samples generated on prompt "A motorized bicycle covered with greens and beans." by SD-v1.5, DDPO, TDPO, UCA, CoCA trained on HPSv2 reward function.}
  \label{step-result-33}

  \centering
  \includegraphics[width=0.9\textwidth]{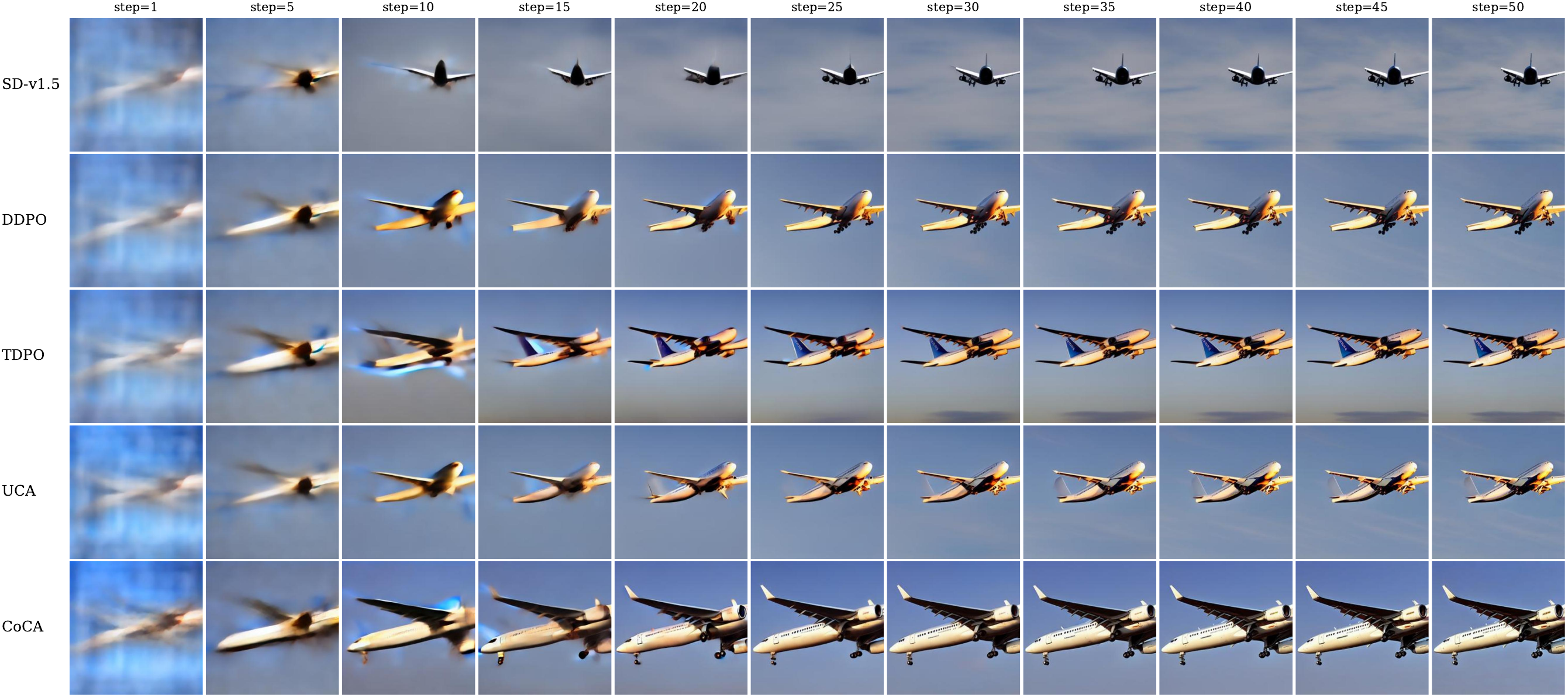}
  \caption{Qualitative comparison of samples generated on prompt "A passenger jet aircraft flying in the sky." by SD-v1.5, DDPO, TDPO, UCA, CoCA trained on HPSv2 reward function.}
  \label{step-result-36}
\end{figure*}

We sample more trajectories of each method to demonstrate rapid global layout changing in trajectories of CoCA, from Figure~\ref{step-result-48} to Figure~\ref{step-result-36}. These samples show that after fine-tuning with CoCA, the denoising process changes faster in the global structure than in the original model or fine-tuning with the trajectory-level reward. This observation further confirms that CoCA reallocates reward to the most influential early steps, enabling the model to form coherent global structures earlier in the trajectory and thereby accelerating convergence during sampling.

